\documentclass[letterpaper, 10 pt, conference]{ieeetrans}  

\IEEEoverridecommandlockouts                              

\usepackage[utf8]{inputenc}
\usepackage{mathtools}
\usepackage{amsmath}
\usepackage{amsfonts}
\usepackage{amssymb}
\usepackage{balance}
\usepackage{lipsum}  
\usepackage{xcolor}
\usepackage{graphicx}
\usepackage{booktabs}
\usepackage{hyperref}
\usepackage[sorting = none, backend = bibtex, style=ieee]{biblatex}
\usepackage{svg}
\usepackage{tabularx} 
\usepackage{subcaption}
\usepackage{dblfloatfix}
\usepackage{array}
\usepackage{caption}
\usepackage{pifont,xcolor}
\usepackage{multirow}

\AtNextBibliography{\small}

\title{\LARGE \bf Decentralized Real-Time Planning for Multi-UAV Cooperative Manipulation via Imitation Learning}

\author{Shantnav Agarwal$^{1}$, Javier Alonso-Mora$^{1}$, and Sihao Sun$^{1}$ 
\thanks{$^1$ Authors are with the Department for Cognitive Robotics, ME, Delft University of Technology, Delft, Netherlands}}

\begin{document}

\maketitle

\begin{abstract}
Existing approaches for transporting and manipulating cable-suspended loads using multiple UAVs along reference trajectories typically rely on either centralized control architectures or reliable inter-agent communication.
In this work, we propose a novel machine learning–based method for decentralized kinodynamic planning that operates effectively under partial observability and without inter-agent communication.
Our method leverages imitation learning to train a decentralized student policy for each UAV by imitating a centralized kinodynamic motion planner with access to privileged global observations.
The student policy generates smooth trajectories using physics-informed neural networks that respect the derivative relationships in motion.
During training, the student policies utilize the full trajectory generated by the teacher policy, leading to improved sample efficiency. Moreover, each student policy can be trained in under two hours on a standard laptop.
We validate our method in both simulation and real-world environments to follow an agile reference trajectory, demonstrating performance comparable to that of centralized approaches.
\end{abstract}
\section{Introduction}
\label{sec:introduction}




Unmanned aerial vehicles (UAVs) have gained significant traction across domains such as surveillance, agriculture, and infrastructure inspection due to their agility and versatility. However, their limited payload capacity restricts their effectiveness in applications involving the transportation of heavy or bulky objects which is common in construction and large-scale logistics. A scalable and cost-effective solution to this limitation is cable-suspended cooperative aerial manipulation \cite{9462539}, where multiple UAVs cooperatively transport and control a cable-suspended payload. This method enables full pose manipulation of objects whose weight may exceed the capacity of a single UAV.

Numerous control strategies have been proposed for cooperative transportation of suspended payloads using UAV teams. These approaches vary in terms of modeling accuracy, scalability, communication requirements, and capability to regulate the full pose of the payload. Given the focus of this work on decentralized cooperative aerial manipulation, prior methods are categorized into three primary frameworks: centralized control, decentralized control with communication, and decentralized control without communication.
\begin{figure}[!ht]
    \centering
    \includegraphics[width=0.8\linewidth]{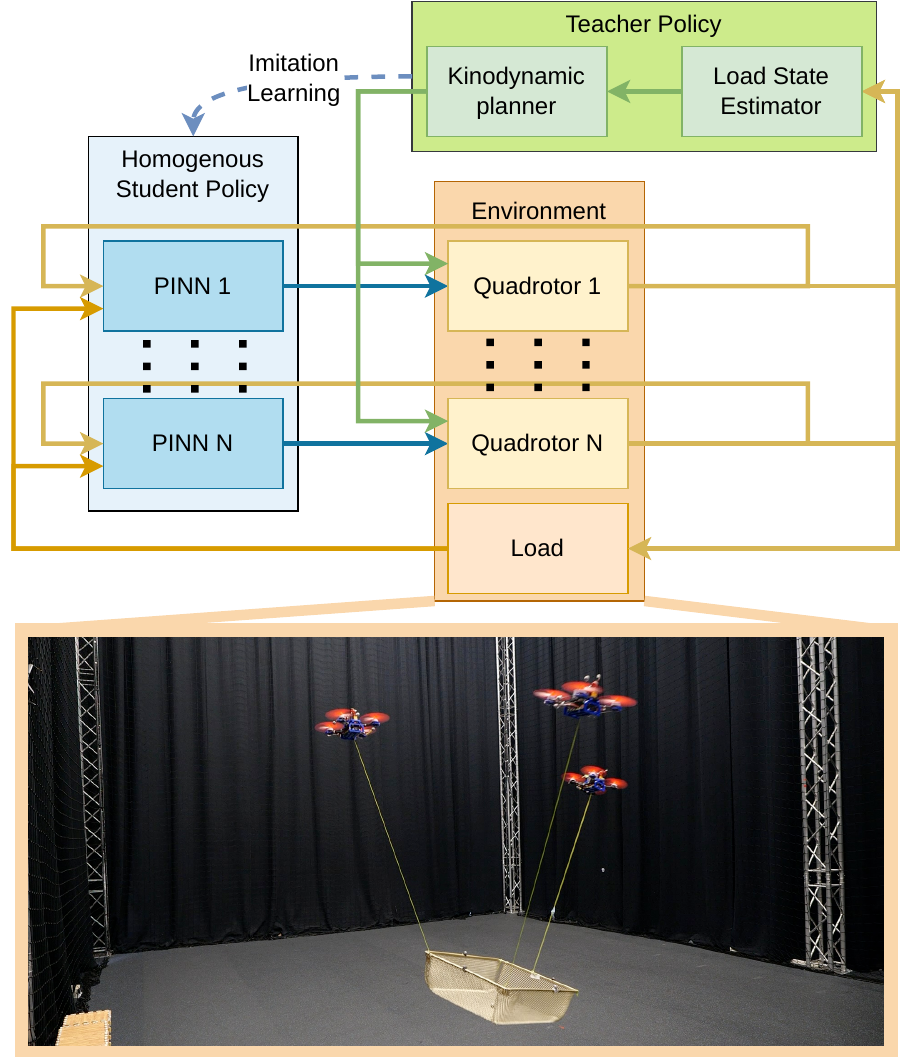}
    \caption{We enable decentralized cooperative aerial manipulation through student policies that operate independently using only the ego UAV’s state and the pose of the load. These student policies are trained via imitation learning from a centralized teacher policy with privileged observations, including the full state of the other UAVs and the load. The policy has been tested in real-world environments, where three UAVs cooperatively manipulate a cable-suspended load.}
    \label{fig:overview}
\vspace{-20pt}
\end{figure}

\subsubsection{Centralized Approaches}
A formation-based geometric control approach was introduced in \cite{6760757}, modeling a point-mass suspended via massless rigid links. This method centrally solves for force equilibrium using full system state information, allowing formation maintenance but neglects payload attitude. Moreover, the approach has only been validated in simulation.

Model Predictive Control (MPC) is commonly used in centralized frameworks due to its ability to handle constraints and optimize control over a receding time horizon. Linear MPC approaches typically model the payload as a point mass and use linearized system dynamics \cite{TARTAGLIONE20171, Moreno_Caireta_2018}. While computationally efficient, these methods cannot control the payload’s orientation and often under perform on dynamic trajectories \cite{kamel2017linear}.

Nonlinear MPC (NMPC) techniques have been introduced that allow manipulating the full-pose of the load along agile trajectories.
For example, a cascaded NMPC controller has been developed that enables full pose manipulation of the load\cite{Li_2023} . However, this controller assumes load dynamics are a magnitude slower than quadrotor dynamics, which reduces control performance on agile trajectories.
Recently, in~\cite{sun2025agilecooperativeaerialmanipulation}, an online kinodynamic motion planner utilizing a load-cable-UAVs model without assumption of time-scale separations has been proposed, enabling agile full-pose control.

While centralized controllers have achieved agile and accurate load manipulation, they rely on high-bandwidth, low-latency communication networks with a central coordinator. This assumption is often impractical in real-world scenarios where communication may be unreliable or unavailable. Further, the computational cost of aforementioned approaches grows exponentially with the number of agents thereby limiting scalability. Hence, the community has also explored decentralized methods for cooperative aerial manipulation.

\subsubsection{Decentralized Approaches with Communication}

A distributed model predictive control framework has been developed which demonstrates performance comparable to centralized MPC while distributing the computational burden across agents\cite{9341541}. The approach assumes a fully-connected communication graph, wherein each UAV shares its previous control inputs with neighboring agents and independently optimizes its own input sequence at each iteration. Though validated only in simulation, the framework exhibits scalability and coordination capabilities. While the method is scalable, it still relies on continuous inter-agent communication, motivating the need for decentralized control strategies that operate without explicit communication.

\subsubsection{Decentralized Approaches without Communication}
A decentralized control scheme combining distance-based formation control with incremental nonlinear dynamic inversion (INDI) is proposed in \cite{8794316}. UAVs compute local acceleration commands based solely on relative position measurements, eliminating the need for global positioning or inter-agent communication. This method, however, does not regulate the payload’s orientation. Master-slave architectures have also been employed, where a designated master UAV generates a reference trajectory, and slave UAVs follow using force-based admittance control \cite{7989678, 8286868}. 
These approaches are limited in their ability to manipulate the payload’s full pose and have not been extended beyond two-UAV systems.
While decentralized approaches without inter-agent communication offer benefits like scalability and robustness, they struggle to achieve agile and precise full-pose manipulation of a payload.
To address this challenge, we present a decentralized, communication-free online kinodynamic motion planner for cooperative aerial manipulation.
Our method enables multiple autonomous UAVs to perform agile and accurate full-pose control of a rigid object.
Each UAV operates independently, relying solely on local observations and the object’s reference trajectory, to generate it's own receding-horizon trajectories.
This trajectory is then followed by low-level trajectory tracking controllers deployed onboard the UAV.
To achieve this, each UAV is equipped with a neural network policy trained through imitation learning from a privileged expert~\cite{DBLP:journals/corr/abs-1912-12294}.
A centralized receding-horizon kinodynamic motion planner serves as the teacher policy~\cite{sun2025agilecooperativeaerialmanipulation}, guiding the training of the strongly homogeneous~\cite{marl-book} decentralized student policies.
We assume that each UAV can accurately estimate both its own state and the state of the payload.
To the best of our knowledge, this is the first work to develop a decentralized, communication-free planner for cooperative cable-suspended aerial manipulation. We have validated our approach in both simulation and real-world experiments.
\section{Problem Statement}
\label{sec:problem_statement}
\subsection{Notations}
We consider a system composed of $n$ unmanned aerial vehicles (UAVs) and a single rigid load body. The pose and twist of the 
$i^{\text{th}}$ UAV are denoted by $[\boldsymbol{p}_i \quad \boldsymbol{q}_i]$ and $[\boldsymbol{v}_i \quad \boldsymbol{\omega}_i]$, respectively, where $\boldsymbol{p}_i \in \mathbb{R}^3$ and $\boldsymbol{q}_i \in \mathbb{S}^3$ represent the position and orientation (quaternion), and $\boldsymbol{v}_i, \boldsymbol{\omega}_i \in \mathbb{R}^3$ denote the linear and angular velocities. We use bold lowercase letters to denote vectors, and bold capitalized letters for matrices; otherwise, variables are scalars.

Similarly, the pose and twist of the load are represented as $[\boldsymbol{p}_L \quad \boldsymbol{q}_L]$ and $[\boldsymbol{v}_L \quad \boldsymbol{\omega}_L]$, with analogous definitions.

A load-attached reference frame, denoted $\mathcal{L}$, is defined such that its origin coincides with $\boldsymbol{p}_L$, and its orientation is aligned with the inertial frame $\mathcal{I}$.

Each UAV has an associated body-fixed frame $\mathcal{B}_i = [\boldsymbol{O}_i \quad \boldsymbol{x}_i \quad \boldsymbol{y}_i \quad \boldsymbol{z}_i]$, where the origin $\boldsymbol{O}_i$ is located at $\boldsymbol{p}_i$, and the $z_i$-axis points vertically upward, coinciding with the thrust direction.

Unless otherwise stated, all vectors are expressed in the inertial frame $\mathcal{I}$. Frame-specific quantities are indicated using superscripts when necessary.

\subsection{Cooperative Aerial Manipulation}

cooperative aerial manipulation refers to the coordinated use of multiple aerial robots to jointly transport and manipulate a common object. A representative depiction of this setup is provided in Figure \ref{fig:overview}. The system consists of multiple UAVs, each connected to a rigid load 
via cables. The cables are attached below the center of mass of each UAV, and at arbitrary points on the load, using universal joints that allow the transmission of forces but not moments.

Each UAV is modeled as a unidirectional thrust platform, capable of generating thrust solely along its body-fixed positive $z_i$-axis. The primary objective is for the UAVs to collaboratively control the full pose of the load—both its position and orientation—so that it tracks a desired trajectory over time.

\subsection{Fully Decentralized Cooperative Aerial Manipulation}


\begin{figure*}[t]
    \centering
    \includegraphics[width=0.9\textwidth]{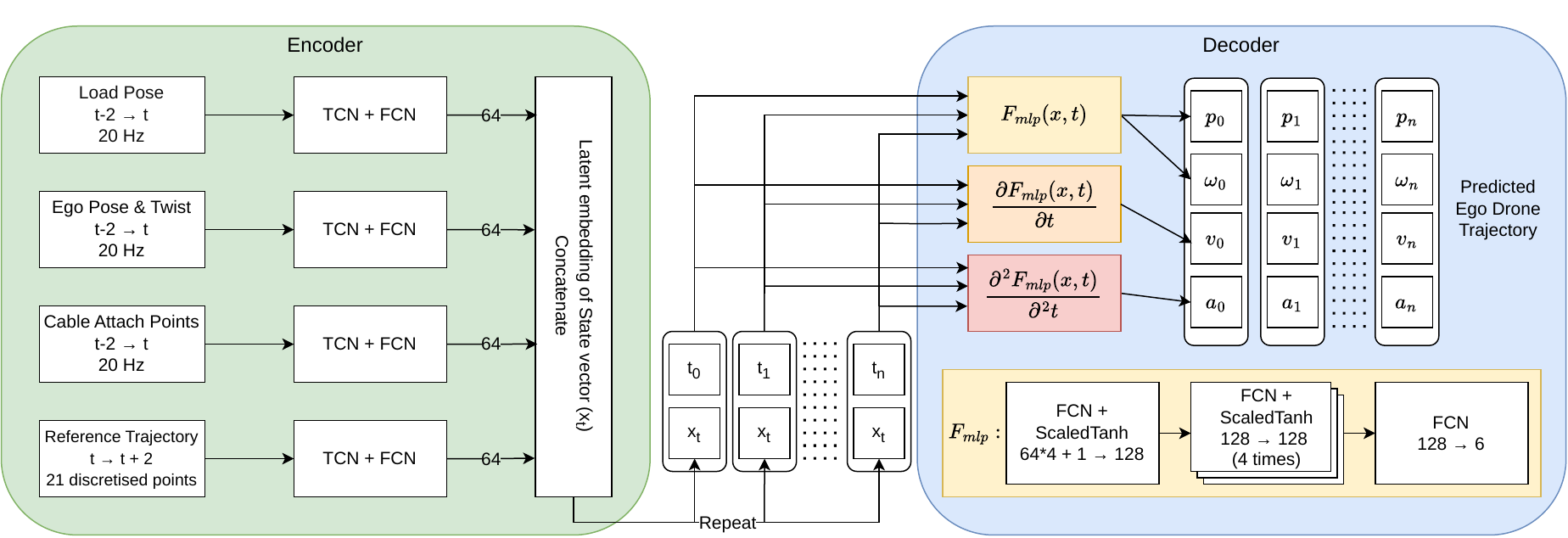}
    \caption{The architecture of the student policy is made up of two parts - Encoder (Sec. \ref{sec:obs_space}) \& Decoder (Sec. \ref{sec:action_space}). The Encoder network maps the observation histories to a latent vector $x_t$. This latent vector is then copied for all nodes in the horizon and passed to the Decoder network.}
    \label{fig:pinn-policy}
\vspace{-20pt}
\end{figure*}

Fully decentralized cooperative aerial manipulation aims to achieve full-pose control of a load using multiple UAVs, where each UAV operates independently based on local observations and without direct communication with other UAVs.

This decentralized setting introduces several key constraints and challenges - 
\textit{Distributed Control:} Each UAV is governed by an onboard control policy and functions autonomously, without access to the control inputs or internal states of other UAVs.
\textit{Local Observations:} Each UAV's planning and control policy relies solely on locally available sensor data, including its own pose and twist, as well as the pose of the load. No information about the states or actions of other UAVs is accessible.
\textit{Implicit Coordination:} UAVs must learn to collaborate effectively during the training phase, such that during deployment, coordinated behavior emerges through physical interactions alone and without any explicit communication channels.
This framework reflects practical limitations in real-world multi-robot systems, where communication constraints, scalability, and robustness to failure are critical considerations.

\section{Method} \label{sec:method}
This section introduces the proposed control policy, including details on the training procedure, observation \& action space as well as the architecture of the learned policy. 

\subsection{Teacher Policy}
We use the planning algorithm introduced in ~\cite{sun2025agilecooperativeaerialmanipulation} as the teacher policy, which is the state-of-the-art centralized method for the multi-lifting system for full-pose control of a cable-suspended load, with high agility and robustness.
Given the reference trajectory of the load, the teacher policy solves an optimization-based kinodynamic motion planning problem and generates receding-horizon reference trajectories for each UAV at 10 Hz.
The reference trajectories are then followed by low-level controllers deployed on each UAV, using a flatness-based controller and INDI~\cite{doi:10.2514/1.G001490} to compensate for external forces and torques from the cable.
\subsection{Student Policy}
While the centralized policy demonstrates high agility and robustness against dynamic model uncertainties, it relies on access to the full state information of all UAVs and the load. In contrast, the student policy learns to generate trajectories for the ego UAV in a decentralized manner, using only the ego UAV's state and the load's pose. Notably, the student policy shares the same low-level controllers as the teacher policy. It is implemented as a strongly homogeneous machine learning policy—meaning all UAVs run identical copies of the policy, with the same architecture and weights~\cite{marl-book}. 
The architecture of the student policy is depicted in Figure \ref{fig:pinn-policy}, and an overview of the overall control framework is illustrated in Figure \ref{fig:overview}. The policy operates at a control frequency of 10 Hz, consistent with the teacher policy.

\subsubsection{Observation Space}
\label{sec:obs_space}
The student policy has access to the following information:
\begin{align}
    \boldsymbol{o}_i&=\big[ \boldsymbol{p}_{L}^{\mathcal{L}} \quad \boldsymbol{q}_{L}^{\mathcal{L}} \quad \boldsymbol{p}_{i}^{\mathcal{L}} \quad \boldsymbol{v}_{i}^{\mathcal{L}} \quad \boldsymbol{q}_{i}^{\mathcal{L}} \quad \boldsymbol{\omega}_{i}^{\mathcal{L}} \quad \boldsymbol{p}_{C1,i}^{\mathcal{L}} \quad  \boldsymbol{p}_{C2,i}^{\mathcal{L}} \big]
    \label{eq:observation_space}
\end{align}
The decentralized policy receives the \textit{Pose} of the load as well as the \textit{Pose} and \textit{Twist} of the ego UAV. 
Notably, the policy does not receive any information about the other UAVs in the team. 

To enable generalization across environments, all observations are expressed in the load frame $\mathcal{L}$. This makes the policy invariant to absolute world position allowing it to operate in spaces of arbitrary size.

It is demonstrated in~\cite{gao2024coohoilearningcooperativehumanobject} that leveraging load dynamics as implicit communication effectively facilitates cooperative load transportation. Their method encodes the load’s state via the positions and velocities of eight bounding-box vertices, where agents’ actions alter these dynamics and influence others’ strategies, enabling scalable multi-agent collaboration.

Similarly, we provide the student policy with the position of the ends of the cable on the load ($\boldsymbol{p}_{C1,i}^{\mathcal{L}}$) and on the UAV 
($\boldsymbol{p}_{C2,i}^{\mathcal{L}}$) The cable attach points, represented in the load-fixed frame $\mathcal{L}$, convey both the spatial position where each UAV applies force and the direction of that force. The exact location of these points on the load defines the moment arm, which determines the torque generated around the load’s center of mass. This information is crucial for accurately controlling the load’s pose during cooperative manipulation.
The positions of these points are computed from known attachment positions in the UAV and load frames:
\begin{equation}
    \boldsymbol{p}_{C1,i} = \boldsymbol{p}_L + \boldsymbol{R}_L \boldsymbol{\rho}_i^\mathcal{L}, \quad 
    \boldsymbol{p}_{C2,i} = \boldsymbol{p}_i + \boldsymbol{R}_i \boldsymbol{r}_i^{\mathcal{B}_i},
\end{equation}
where $\boldsymbol{R}_L$, $\boldsymbol{R}_i$ is the rotation matrix of the load and the UAV, respectively. $\boldsymbol{\rho}_i^\mathcal{L}$ represents the vector from load origin to the $i^{th}$ UAV's cable attach point on the load. Similarly, $\boldsymbol{r}_i^{\mathcal{B}_i}$ represents the vector from the the ego UAV's origin to the cable attach point.

Instead of using quaternions, we represent rotations using the 6D continuous formulation proposed by Zhou et al. \cite{8953486}. This representation avoids gimbal lock and discontinuities inherent in Euler angles or quaternions and therefore improves differentiability and learning stability.

To reason under partial observability and infer the intentions from other agents, the policy uses a history of past observations. Specifically the observation matrix at time $t$ is defined as
\begin{equation}
    \boldsymbol{O}_i = \left[ \boldsymbol{o}_{i,t}, \boldsymbol{o}_{i, t - dt}, \dots, \boldsymbol{o}_{i, t - N dt} \right]
\end{equation}
where $\boldsymbol{o}_{i,t}$ is the observation vector of the $i$-th UAV at time $t$.

Typically, time parameterized representations of future reference states help the agent develop a deeper understanding of the scenario, leading to improved performance \cite{DBLP:journals/corr/abs-2107-09647}. Therefore, we provide the policy with the reference trajectory for the next 2 seconds at $N = 21$ nodes, which are the same references for the teacher policy, thereby ensuring that both the teacher and student get the same reference data as input during training.
\begin{equation} \label{eqn:y_ref}
    \boldsymbol{Y} = \left[ \boldsymbol{y}_0 \quad \boldsymbol{y}_1 \dots \boldsymbol{y}_N \right]
\end{equation}
where 
\begin{equation}
    \boldsymbol{y}_j =  \left[ \boldsymbol{p}_{L}^\mathcal{L} \quad \boldsymbol{v}_{L}^\mathcal{L} \quad \boldsymbol{a}_{L}^\mathcal{L} \quad \boldsymbol{q}_{L}^\mathcal{L} \quad \boldsymbol{\omega}_{L}^\mathcal{L} \right]_j,~j \in \{0, \dots, N\}.
\end{equation}
Here $\boldsymbol{a}_{L}$ represents the linear acceleration of the load. 

To process sequential data of observation histories and reference trajectories, we use Temporal Convolutional Networks (TCN), which offer longer memory, stable gradients, and fast, parallel inference. Then the outputs of the TCN network are passed through a dense layer to create a vector of latent embeddings.
The four streams of information as shown in Figure \ref{fig:pinn-policy} are recorded at different time steps and therefore processed using separate TCN networks. The outputs of all TCNs are concatenated into a latent vector $\boldsymbol{x}_i$ that encodes the current state of the system as well as the desired future reference states:
\begin{equation}
    \boldsymbol{x}_i = \pi_{\text{encoder}}(\boldsymbol{O}_i, \quad \boldsymbol{Y})
\end{equation}


\subsubsection{Action Space}
\label{sec:action_space}
The student policy, as an online planner, directly generates reference trajectories for the ego UAV.
In the training time, we use the output of the centralized teacher policy's online planner directly as the target for the student policy.
Therefore, the action space for the $i$-th UAV can be represented by the following equations. Note that we omit the subscript $i$ for readability.

\begin{equation} \label{eqn:action_space}
    \boldsymbol{U} = \left[ \boldsymbol{u}_0 \quad \boldsymbol{u}_1 \dots \boldsymbol{u}_N \right]
\end{equation}
with
\begin{equation} \label{eqn:action_space_2}
    \boldsymbol{u}_j = \left[ \boldsymbol{p} \quad \boldsymbol{\omega} \quad \boldsymbol{v} \quad \boldsymbol{a} \right]_j ,\quad j \in \{0, \dots, N\}
\end{equation}
where $\boldsymbol{p}$, $\boldsymbol{v}$, $\boldsymbol{a}$ represent desired position, velocity and acceleration; $\boldsymbol{\omega}$ represents the desired body rates of the ego UAV.
The onboard low-level controller typically employs a differential-flatness-based method to follow the reference trajectory generated by the planner for the UAV. Incorporating higher-order derivatives of the reference trajectory, such as velocity and acceleration, can significantly improve tracking performance \cite{sun2022comparative}.
A key challenge in predicting higher-order terms with a learning-based policy is ensuring kinematic consistency. The network may generate trajectories wherein the time derivative of position deviates from the predicted velocity, and likewise, the derivative of velocity from the predicted acceleration.
To address this problem, we propose the use of Physics Informed Neural Networks (PINNs) with architectural constraints applied as described below:


\begin{equation}
    [\boldsymbol{p}_j, \quad \boldsymbol{\omega}_j] = F_{mlp}(\boldsymbol{x}_i, t_j)
\end{equation}
\begin{equation}
    \boldsymbol{v}_j = \frac{\partial F_{mlp}(\boldsymbol{x}_i, t_j)}{\partial t}
\end{equation}
\begin{equation}
    \boldsymbol{a}_j = \frac{\partial^2 F_{mlp}(\boldsymbol{x}_i, t_j)}{\partial t^2}
\end{equation}
where t is the instantaneous time and $t_j$ is the timestamp of the $j$-th node in the UAV reference trajectory calculated using the equation:
\begin{equation}
    t_j = t_{j-1} + 0.01 + 0.009 \cdot (j-1), \quad t_0 = t
\end{equation}

We therefore employ PINNs as the decoder (Fig.~\ref{fig:pinn-policy}) to generate trajectories from latent vectors that are inherently feasible, as they explicitly enforce kinematic constraints, i.e., that velocity is the derivative of position and acceleration is the derivative of velocity. 

\subsection{Training environment} \label{sec:training_env}
The student policy is exclusively trained within the Gazebo (RotorS) \cite{Furrer2016} simulation environment, and the resulting policy is subsequently deployed in the real world without any modification. The training process employs privileged learning \cite{DBLP:journals/corr/abs-1912-12294} in conjunction with the Dataset Aggregation (DAgger) algorithm \cite{ross2011reductionimitationlearningstructured}.
DAgger is an iterative imitation learning algorithm wherein each iteration consists of two primary phases: data collection and policy training. During data collection, either the teacher or the student policy is used to interact with the environment to collect demonstrations. Afterwards the policy is trained to predict the teacher's action given the corresponding observations as input. In the initial iteration, only the teacher policy is used to determine actions. In subsequent iterations, the probability of using the student policy to control the system is gradually increased. This approach enables the generation of a more diverse dataset that encompasses a wider range of system states, thereby enhancing the robustness of the learned student policy \cite{ross2011reductionimitationlearningstructured}.

To further improve the quality and diversity of the training dataset, we introduce randomized external disturbances during the data collection phase. Specifically, at the start of an episode, a random force-torque vector is selected to define the disturbance direction. During each time step of the episode, a force with random magnitude is applied along this predefined direction, effectively simulating a consistent yet stochastic perturbation. This method promotes the exploration of a broader range of system dynamics, which contributes to the robustness and generalization capability of the resulting policy.

\section{Results}
\label{sec:results}
\subsection{Environment for practical experiments}
The proposed algorithm is validated through real-world experiments involving cooperative manipulation of a 1.4 kg rigid-body payload using three quadrotors, each with a mass of 0.6 kg. The payload is suspended by three cables, each 1 m in length, connected to distinct attachment points on the payload to facilitate full pose control. The opposing ends of the cables are affixed 0.03 m below the center of gravity (CoG) of each quadrotor. The aerial platforms are customized from the Agilicious open-source hardware framework \cite{doi:10.1126/scirobotics.abl6259}, with onboard computation for low-level controllers handled by Raspberry Pi 5 mini PCs.

The student policy for each UAV is deployed as an independent ROS node on the aforementioned laptop. The predicted trajectories are transmitted to the Agilicious flight controller onboard each UAV via a ROS topic over Wi-Fi at 10 Hz. We use a motion capture system for high-precision state estimation of both the UAVs and the suspended payload.

\begin{figure}[b!]
    \centering
    \includegraphics[width=0.8\linewidth]{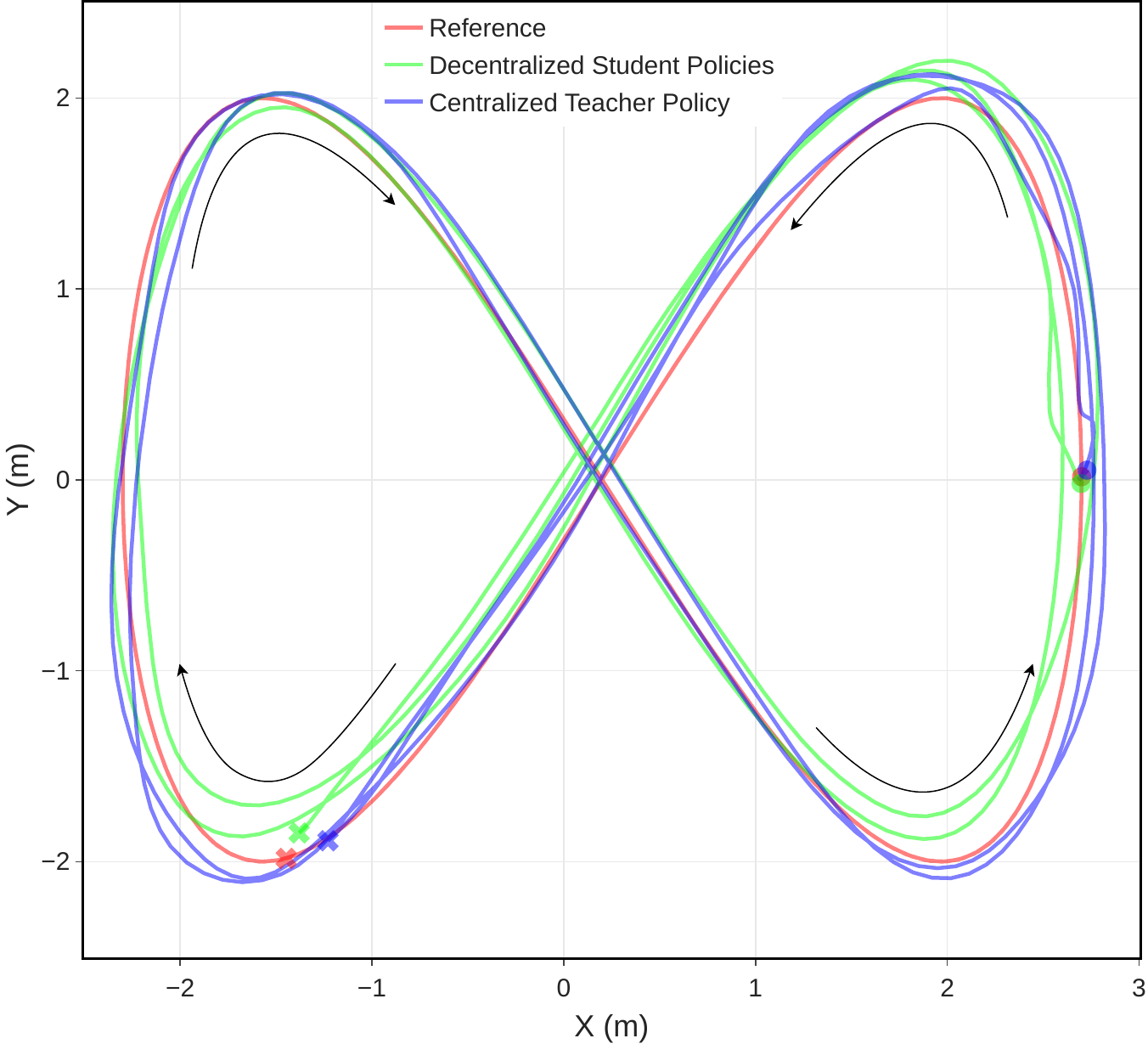}
    \caption{Top view of tracking performance. Circle, cross indicate start, end of trajectory. Two laps of Eight are performed.}
    \label{fig:tracking_perf}
\end{figure}
\subsection{Pose Trajectory Tracking}
The student policy was trained on a figure-eight trajectory with dimensions 2.2 m $\times$ 2 m over 20 DAgger rounds, each consisting of a single episode. After each round, the probability of selecting actions from the student policy was linearly increased from 0 to 1 by round 20. During each round, the policy was trained on the aggregated dataset for 16 epochs. To enhance robustness against real-world perturbations, consistent yet stochastic disturbances, as detailed in Section \ref{sec:training_env}, were applied to the load during the final 5 episodes of training. The entire training process, including data collection and policy updates, was executed on the same laptop used for deployment and completed within 80 minutes. The trained policy was then evaluated on the identical figure-eight trajectory in the real world via zero-shot sim-to-real transfer. We perform two flights for both student and teacher policies on the same hardware and report the average of the metrics in Table \ref{tab:tracking_perf}. We also show the trajectory tracking performance of both in Fig \ref{fig:tracking_perf}.
In real-world experiments, the student policy, executed in a fully decentralized manner using only ego-UAV observations, achieves a comparable orientation RMSE to the teacher policy but exhibits a higher position RMSE. 

Next, we train the student policy in a similar manner, but using two distinct trajectories. Training alternated between these two trajectories over 20 DAgger rounds, with perturbations applied during the final 4 rounds. The resulting policy was then evaluated in simulation on multiple unseen trajectories. The results are summarized in Table \ref{tab:figure8_split}.

\begin{table}[h]
\centering
\captionsetup{font=small}
\scriptsize
\renewcommand{\arraystretch}{1.2}
\begin{tabular}{|c|c|c|}
\cline{1-3}
\textbf{Metric}        & \textbf{Teacher Policy} & \textbf{Student Policies} \\
\cline{1-3}
RMSE Posn. (m)   & 0.171      & 0.377   \\
RMSE Orient. ($^\circ$)  & 10.447   & 9.335 \\
\cline{1-3}
\end{tabular}
\caption{Comparison of metrics from real-world flights following the figure-eight trajectory shown in Fig~\ref{fig:tracking_perf}. The values are the means of the two flights.}
\label{tab:tracking_perf}
\vspace{-20pt}
\end{table}

\begin{table}[h]
\captionsetup{font=small}
\centering
\scriptsize
\renewcommand{\arraystretch}{1.2}
\begin{tabular}{|c|c|c|cc|cc|}
\cline{1-7}
\multicolumn{2}{|c|}{\textbf{Ref. Size}} & \textbf{Train} & \multicolumn{2}{c|}{\textbf{Student Policies}} & \multicolumn{2}{c|}{\textbf{Teacher Policy}} \\
\cline{1-2} \cline{4-7}
\textbf{X(m)} & \textbf{Y(m)} & \textbf{Dataset} & Posn. m & Orient. $^\circ$ & Posn. m & Orient. $^\circ$ \\
\cline{1-7}
2.7 & 2.7 & Yes & 0.166 & 6.768  & 0.089 & 4.111 \\
2.7 & 2.3 & Yes & 0.155 & 5.684  & 0.091 & 4.324 \\
2.0 & 2.0 & No  & 0.377 & 10.877 & 0.073 & 3.397 \\
3.0 & 3.0 & No  & 0.175 & 7.531  & 0.105 & 5.371 \\
3.5 & 3.5 & No  & 0.707 & 12.013 & 0.192 & 6.912 \\
2.5 & 2.5 & No  & 0.205 & 6.903  & 0.090 & 4.196 \\
\cline{1-7}
\end{tabular}
\caption{Comparison of the student and teacher policy 
performance in simulation for multiple trajectories. The table reports the RMSE for position and orientation. Student policies were trained using the first two trajectories. X and Y denote the amplitudes of the sine and cosine functions used to generate the trajectories.}
\label{tab:figure8_split}
\vspace{-20pt}
\end{table}

\subsection{Advantage of PINN over MLP as Decoder}A benefit of using PINNs as decoder is that the trajectories are guaranteed to be consistent and smooth. Consistency comes from the fact that in all target trajectories in training data start at ego UAV's pose and therefore the policies learn this behavior as well. Smoothness are guaranteed by the partial derivative constraints on velocity and acceleration. We compare the PINN architecture with a baseline policy where the \textit{Decoder} network employs a single Multi-layer Perceptron (MLP) to generate all four trajectory components. The trajectories produced by this model are visibly jagged (shown in the supplementary video), lacking the smoothness enforced by the PINN’s constraints. Moreover, they exhibit inconsistencies, where the numerical derivative of predicted position does not match the predicted velocity as highlighted in Table \ref{tab:pinn_ml_table}


\begin{table}[ht]
\captionsetup{font=small}
\centering
\scriptsize
\renewcommand{\arraystretch}{1.2}
\begin{tabular}{|l|c|c|c|c|}
\cline{1-5}
\textbf{Method} & \textbf{X (m/s)} & \textbf{Y (m/s)} & \textbf{Z (m/s)} & \textbf{Overall (m/s)} \\
\cline{1-5}
PINN & 0.010 & 0.030 & 0.004 & 0.015 \\
MLP   & 0.080 & 0.116 & 0.048 & 0.081 \\
Teacher & 0.012 & 0.033 & 0.003 & 0.016 \\
\cline{1-5}
\end{tabular}
\caption{The Mean Absolute Error (MAE) between the numerical derivative of predicted position and the predicted velocity was evaluated for each method. For PINN and the teacher planner, the errors primarily arise from numerical differentiation. In contrast, the MLP exhibits higher errors due to both numerical differentiation and additional inaccuracies stemming from infeasible position and velocity predictions.}
\label{tab:pinn_ml_table}
\vspace{-20pt}
\end{table}

\subsection{Generalizability to Unseen Trajectories}
We train the student policy to execute different trajectories—namely, Eight, Circle, and Square each with varying sizes. These set of trajectories contain a very diverse set of curves with a large range of radii of curvature which needs to be followed by the load. Further, we train the student policy to follow two different orientation modes: 1) Constant orientation where the orientation of the load is constant; and 2) Zero Side-slip orientation where the orientation of the load changes according to the velocity vector.

The trained student policy is then evaluated on the Zandvoort F1 Track located in The Netherlands.
This trajectory differs substantially from those encountered during training, exhibiting greater variability in curvature radii, path length, and overall duration. The quantitative results are summarized in Table \ref{tab:zandvoort_perf}. Additional details on training trajectories and results are provided in Appendix.

\begin{table}[h]
    \centering
    \begin{tabular}{|c|c|c|}
    \cline{1-3}
    \textbf{Metric}        & \textbf{Zandvoort (Constant)} & \textbf{Zandvoort (ZSS)} \\
    \cline{1-3}
    RMSE Posn. (m)   & 0.092      & 0.378   \\
    RMSE Orient. ($^\circ$)  & 3.157   & 93.575 \\
    \cline{1-3}
    \end{tabular}
    \caption{Comparison of metrics from simulated flights around the Zandvoort F1 Track. The Student Policy (Both) was deployed for this experiment without any fine-tuning.}
    \label{tab:zandvoort_perf}
    \vspace{-20pt}
\end{table}

\subsection{Computational Time}
Real-time inference is possible on the Raspberry Pi 5 CPU and takes 97 milliseconds, as the planner generates trajectories at 10~Hz. However, during experiments, we realized that running both the Agilicious flight control stack and PINN inference on the Raspberry Pi leads to a degradation in the inner control loop update rate. Therefore, we eventually deployed PINNs on a laptop and sent them to each UAV through Wi-Fi. 
The average inference time of the student policy is 27 milliseconds, on a NVIDIA RTX 3060 Laptop GPU.
In comparison, the teacher policy requires an average computation time of 15.3 milliseconds (for 3 UAVs) running on an Intel Core i7-13700H CPU according to~\cite{sun2025agilecooperativeaerialmanipulation}, and its computational complexity increases exponentially with the number of UAVs. 
In contrast, the student policy, due to its decentralized and partially observable architecture, exhibits the potential for scalability. By enabling each UAV to independently predict its own trajectory, the framework could support horizontal scaling, making it feasible to accommodate a large number of UAVs without any increase in computation time.
\section{Discussions}
\label{eq:discussions}
The proposed student policy demonstrates the capability to control the full pose of a load along agile reference trajectories without relying on networked communication. Trained entirely in simulation, it achieves zero-shot sim-to-real transfer thanks to the onboard-deployed trajectory tracking controller. Despite being a decentralized machine learning policy, it retains several key benefits of the teacher policy: \textit{Sample Efficiency}, as the entire prediction horizon is leveraged during training; \textit{System Dynamics}, since the learned trajectories are physically consistent by design due to architectural constraints; \textit{Receding Horizon Planning}, which can increase the situational awareness of human operators and aid the development of safety modules; and \textit{Lower Frequency}, as the trajectory-based structure requires execution at lower update rates, allowing for larger models and reduced computational load.

\begin{figure}
    \centering
    \includegraphics[width=0.9\linewidth]{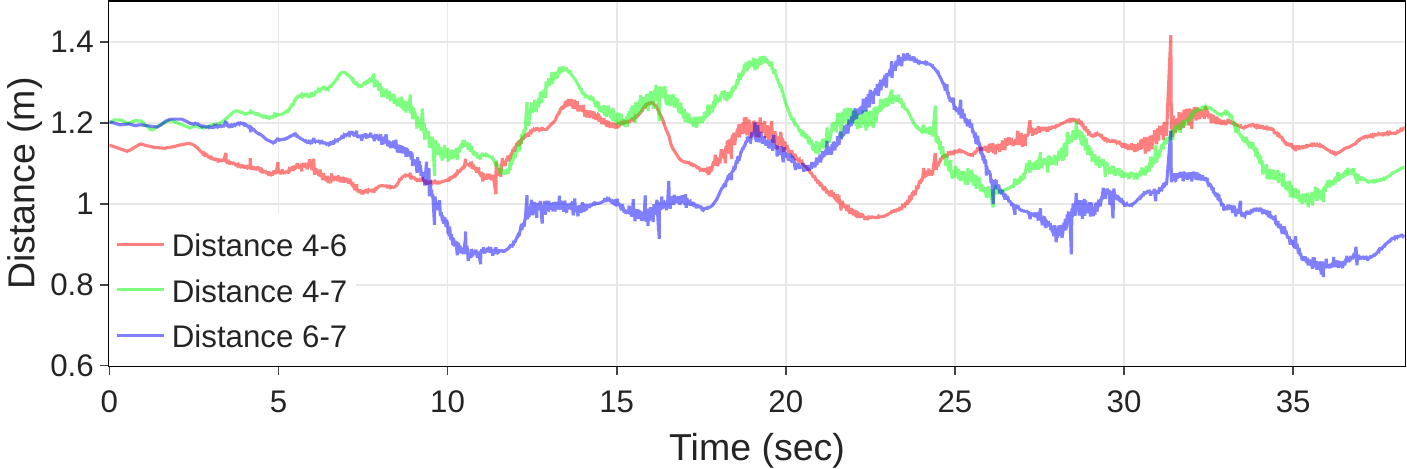}
    \caption{Despite lacking awareness of each other's positions, the quadrotors successfully maintain safe inter-agent distances}
    \label{fig:drone_dist}
\vspace{-20pt}
\end{figure}
One notable limitation of the student policies is the absence of hard guarantees for inter-agent collision avoidance. Unlike the teacher policy, which explicitly enforces minimum separation constraints, the decentralized policy employed in our method lacks direct awareness of other UAVs’ positions. Consequently, the risk of inter-agent collisions is inherently higher for the decentralized student policy. While the student policy learns to approximate safe behavior, primarily by imitating the teacher and placing UAVs in safe configurations relative to the current load pose and reference trajectory, it does not explicitly account for neighboring agents. In real-world experiments, occasional breaches of the 1 m minimum separation threshold were observed; however, the UAVs typically recovered and reestablished safe distances, as illustrated by the trajectory data in Fig. \ref{fig:drone_dist}. 
Enhancing the system with onboard sensing for inter-UAV perception and obstacle detection represents a promising avenue for future research, potentially enabling safer decentralized coordination in aerial manipulation scenarios.

Simulation results on Zandvoort F1 Track indicate that the student policies are able generalize to very different trajectories when the orientation of the load is kept constant. 
However, the pose-tracking performance deteriorates for unseen trajectories with zero side-slip orientation. In these cases, the student policy struggles to compute optimal trajectories, likely because the required motion patterns deviate substantially from those encountered during training. A more comprehensive analysis of the method’s ability to handle a wider range of trajectory types remains an open research direction for future work.

\section{Conclusion}
\label{sec:conclusion}

In this work, we have designed and evaluated a fully decentralized real-time planning method for cooperative aerial manipulation of a cable-suspended load using multiple UAVs, without inter-UAV communications.
Our method leverages imitation learning to train a student policy by imitating a centralized teacher policy with privileged information of the environment. 
We have experimentally validated the proposed planning method, showing its effectiveness in real-world flights.


\printbibliography
\clearpage
\section*{Appendix}
\subsection{Inter-agent Distance}
\begin{figure*}[htbp]
    \centering

    \begin{subfigure}[b]{0.4\textwidth}
        \centering
        \includegraphics[width=\linewidth]{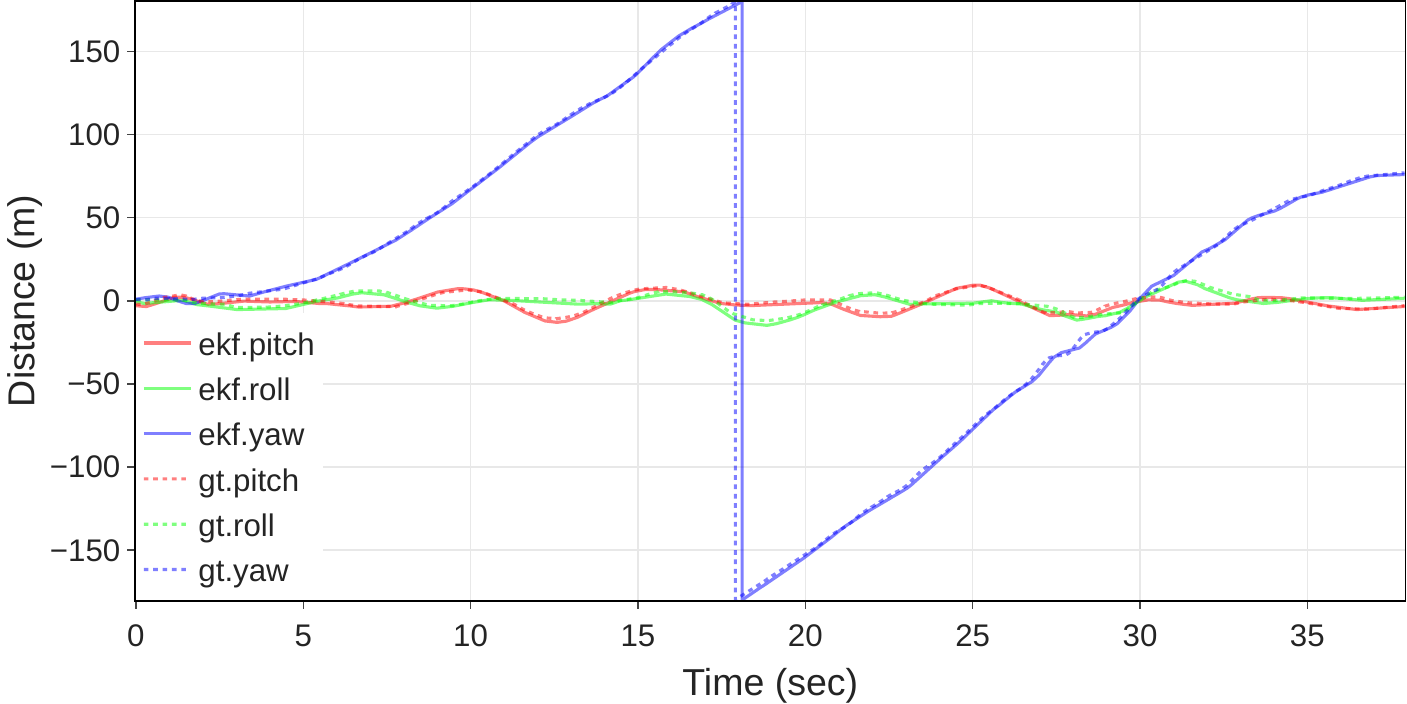}
        \caption{Simulation: Load pose estimation (solid) and ground truth (dot) in simulation environment. In simulation, the EKF works well, and the estimations are accurate.}
        \label{fig:pinnv1_sim_ekf_est}
    \end{subfigure}
    \hfill
    \begin{subfigure}[b]{0.4\textwidth}
        \centering
        \includegraphics[width=\linewidth]{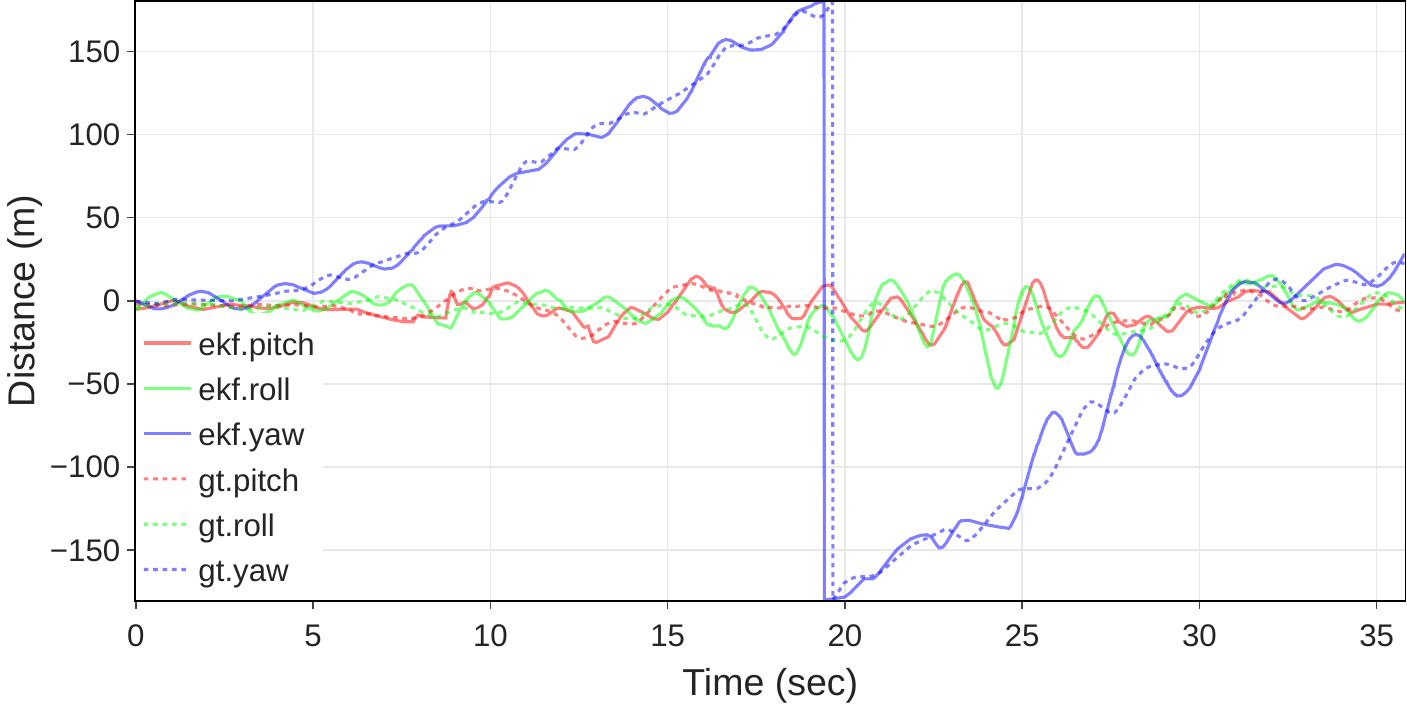}
        \caption{Real World: Pose estimate oscillates around the ground truth pose. The student policies are not robust to this noise in load pose estimation.}
        \label{fig:pinnv1_real_ekf_est}
    \end{subfigure}

    \begin{subfigure}[b]{0.4\textwidth}
        \centering
        \includegraphics[width=\linewidth]{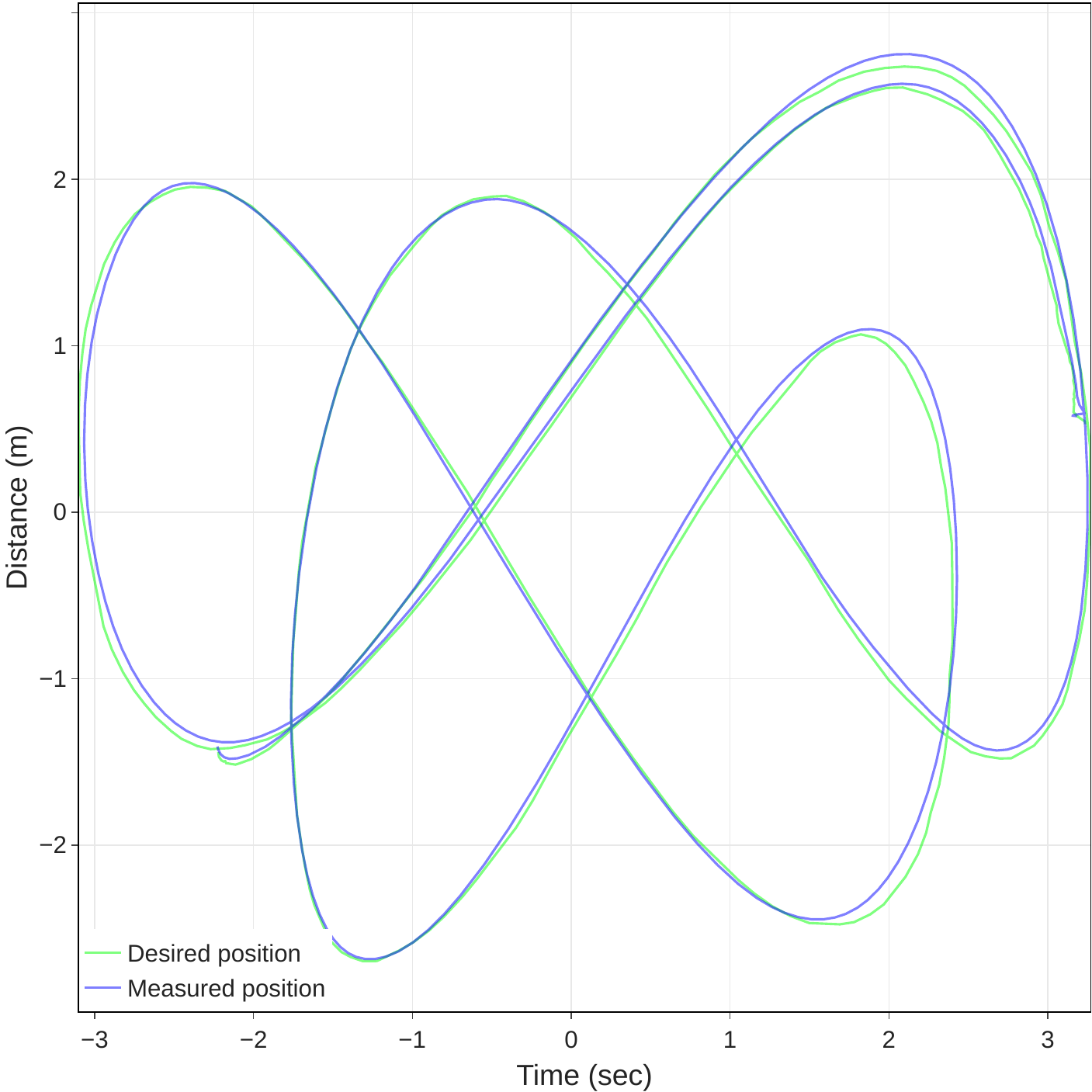}
        \caption{Student Policy EKF in Simulation: Desired and measured position of the UAV is shown. Desired position is the first point of trajectory generated at each time step. Predictions are consistent and can be tracked by the onboard controller.}
        \label{fig:pinnv1_sim_strat_state}
    \end{subfigure}
    \hfill
    \begin{subfigure}[b]{0.4\textwidth}
        \centering
        \includegraphics[width=\linewidth]{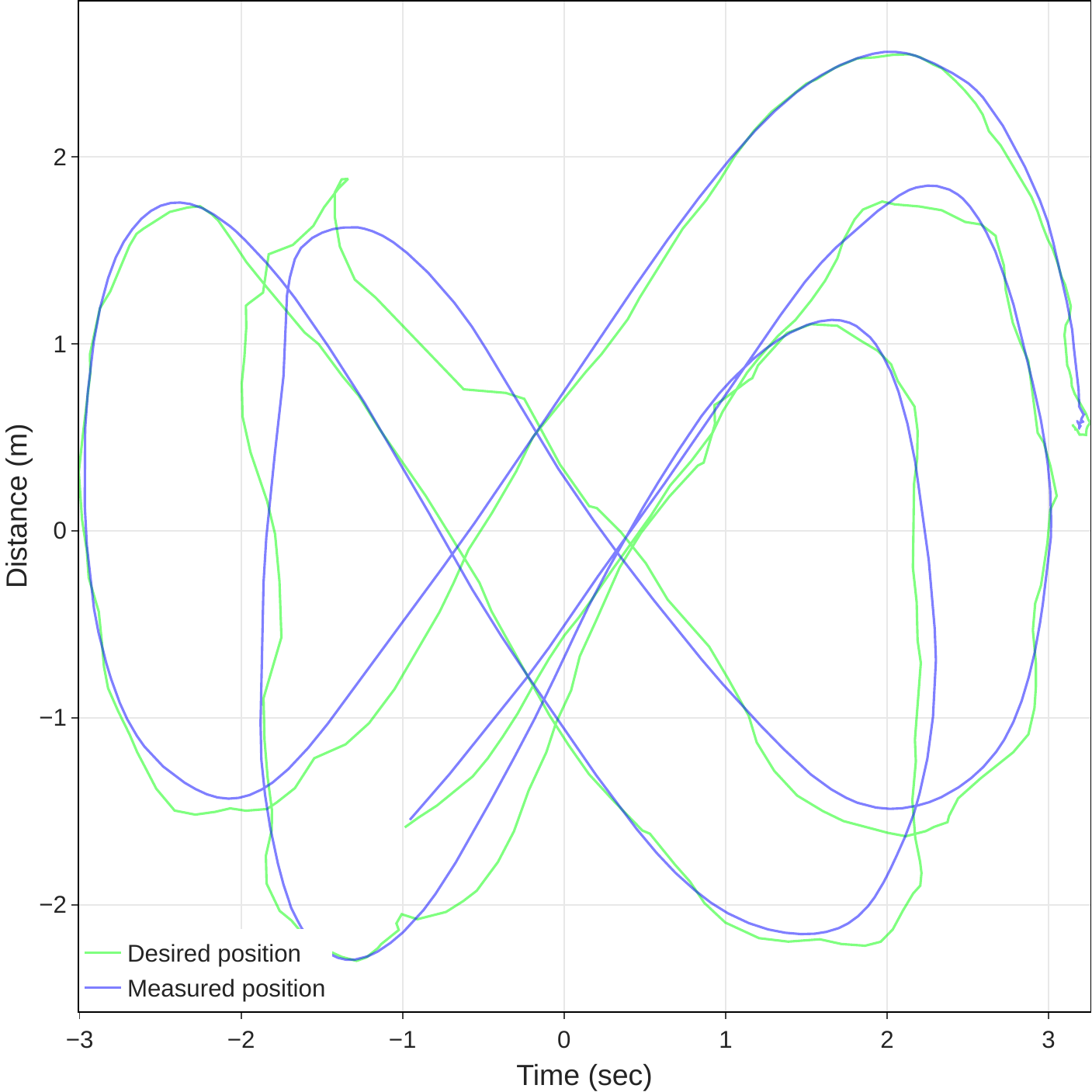}
        \caption{Student Policy EKF in Real World: Policies depend on Load pose information to place UAVs in safe positions. Oscillating Load pose estimations cause the desired positions to be inconsistencies which UAVs cannot track.}
        \label{fig:pinnv1_real_strat_state}
    \end{subfigure}

    \begin{subfigure}[b]{0.4\textwidth}
        \centering
        \includegraphics[width=\linewidth]{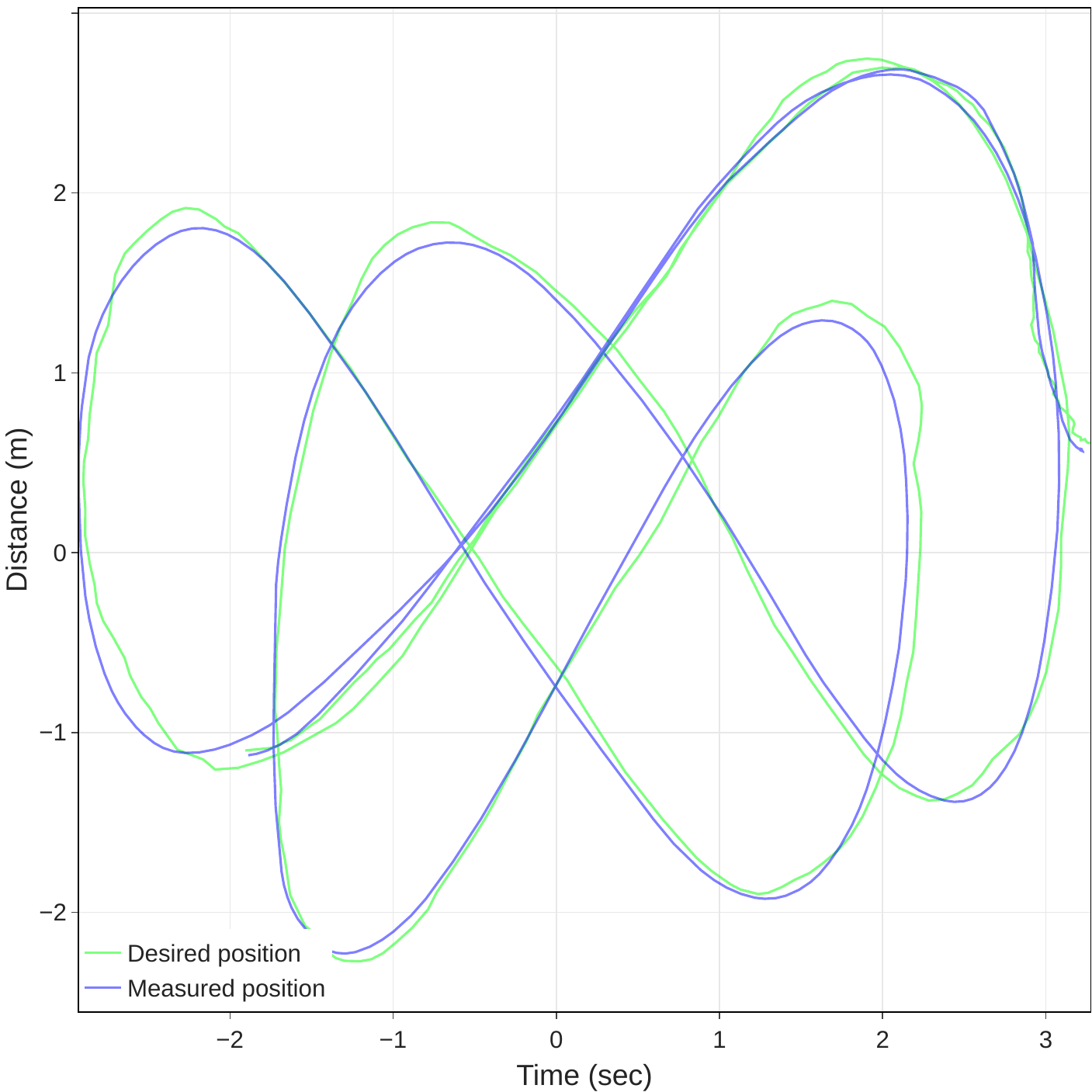}
        \caption{Student Policy MoCap in Real World: Second version of the student policy using ground truth load pose. The trajectory generated by this policy is again consistent and robust to real world perturbations.}
        \label{fig:pinnv2_real_strat_state}
    \end{subfigure}

    \caption{Left: Performance of the system in simulation | Right: Performance of the system when deployed in the real world | Bottom: Performance of the second version of student policies in real world.}
    \label{fig:ekf_strat_plots}
\end{figure*}

One notable limitation of the student policies is the absence of hard guarantees for inter-agent collision avoidance. The teacher policy explicitly enforces minimum separation constraint of 1 m between all drones using a hard constraint in its optimal policy formulation. However, it is not possible to program such constraints in the student policies as they lack direct awareness of other UAV's positions. In the absence of hard guarantees for maintaining minimum separation, the risk of inter-agent collisions is inherently higher for the decentralized student policy.

The student policy is expected to use the load pose and desired reference trajectory to place UAVs in a safe configuration. Getting the student policies to learn these safe configurations was difficult and required some changes to the student policy and the training method. We describe this in detail in the following subsections:

\begin{figure}[htbp]
    \centering
    \begin{subfigure}[b]{0.48\textwidth}
        \centering
        \includegraphics[width=\linewidth]{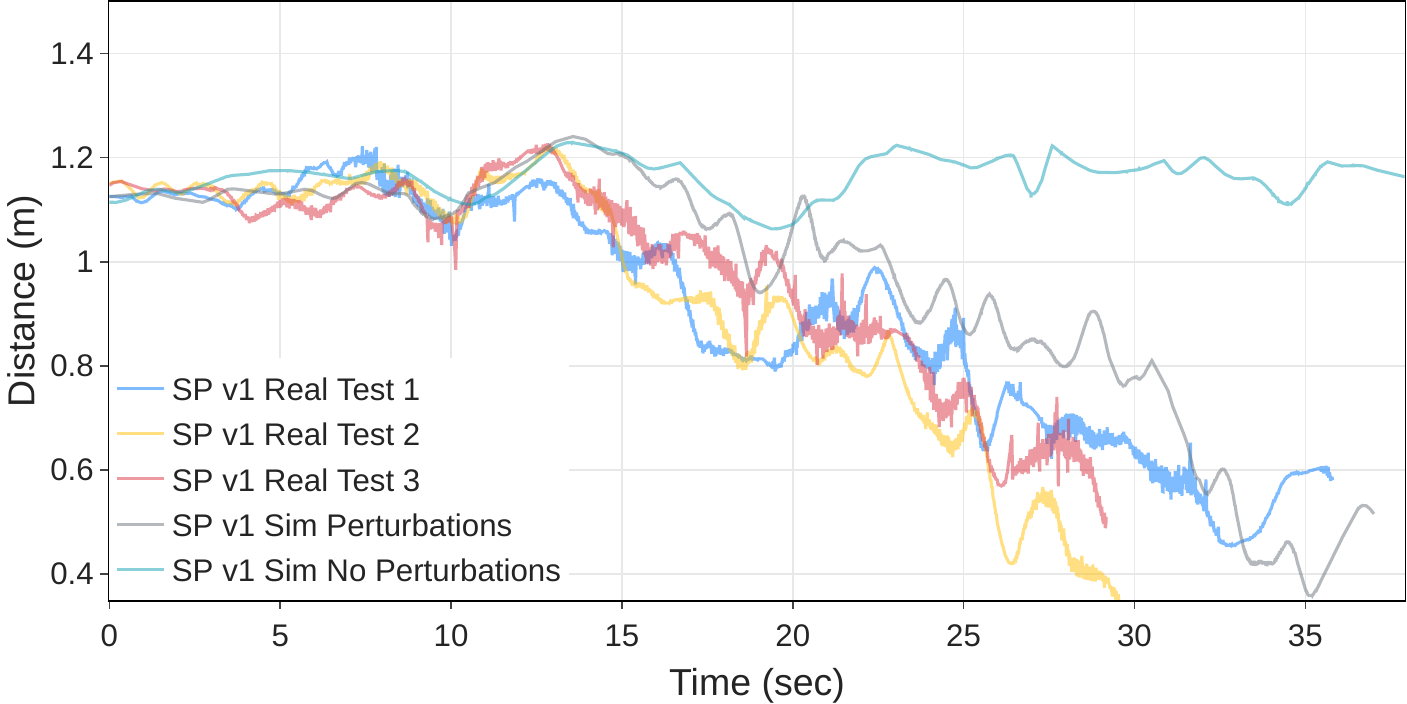}
        \caption{Student Policy EKF: This version of student policies is unable to maintain a safe separation between agents when external disturbances are present.\\~}
        \label{fig:pinn_v1_drone_dist}
    \end{subfigure}

    \begin{subfigure}[b]{0.48\textwidth}
        \centering
        \includegraphics[width=\linewidth]{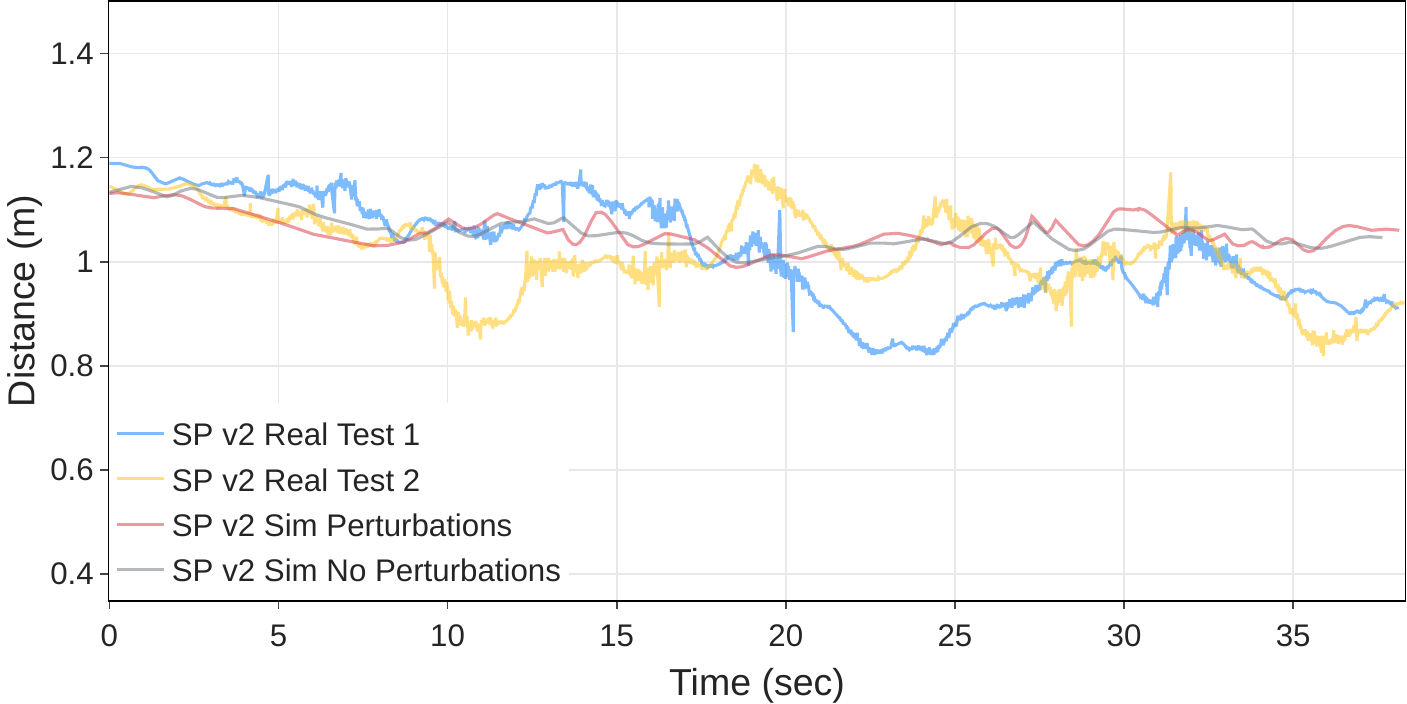}
        \caption{Student Policy MoCap: Despite lacking awareness of each other's positions, the quadrotors successfully maintain safe distances despite the domain gap between the real world and simulation.}
        \label{fig:drone_dist2}
    \end{subfigure}
    \caption{Inter-agent distance for various experiments in simulation and the real world. Minimum inter-agent distance is plotted.}
    \label{fig:all_drone_dist}
\end{figure}

\subsubsection{Load state estimation using an extended Kalman filter}
The teacher policy utilizes an extended Kalman filter to determine the pose of the load. This filter uses the instantaneous pose of all UAVs along with the known attachment points of the cables to estimate the cable directions and load pose, twist. In the first version of the student policy (\textit{Student Policy EKF}), we utilized the values of this Kalman filter as the input for the load state. This version took as input the twist of the load state in addition to other observations mentioned in Equation \ref{eq:observation_space}. While the use of the EKF filter made a part of our solution centralized - the EKF required instantaneous pose of all UAVs; it removed the requirement for measuring the pose of the load - making it more practical for real-world deployment.

This version of the student policies worked well in simulation, however it performed poorly in real world experiments. In real world, the UAVs would progressively come closer to each other as the trajectory progressed. Once the UAVs breached the 0.4 m minimum safe separation between the UAVs, the safety module (described in Section \ref{sec:safety_module}) would intervene and turn off the UAVs to prevent mid-air collisions. This behaviour was not observed in the simulation environment where the UAVs kept a safe distance from each other. We could induce this behaviour in simulation by applying random directed moments on the load.

Upon analysing the data collected during real-world experiments, we identified that the primary source of performance degradation was noisy state estimation from the Extended Kalman Filter (EKF). While the EKF provided accurate load orientation estimates in simulation (Figure \ref{fig:pinnv1_sim_ekf_est}), its performance deteriorated significantly in the real-world deployment due to imperfect UAV state estimation and network-induced delays, resulting in high-frequency noise in the orientation estimates (Figure \ref{fig:pinnv1_real_ekf_est}).
The student policies did not encounter such noisy measurements during training and were therefore unable to cope with them effectively in real-world deployment.
Consequently, the student policy struggled to maintain safe positioning of the ego UAV relative to the load and other UAVs.
Additionally, the noise in load state estimation led to instability in the predicted strategies (Figure \ref{fig:pinnv1_real_strat_state}), causing abrupt changes that the low-level onboard controllers could not reliably track.

\subsubsection{Load state estimation using motion capture}
Noisy load pose estimation was attributed to be the main cause of the failure of student policies in the real world. We therefore, changed the architecture of the student policies to use the ground truth pose values estimated using the motion capture system. During training, this ground truth information is provided by the Gazebo simulator. We further add random wrenches on the load during training to make the policies robust and learn to maintain safe inter-agent distances. This second version of the student policy (\textit{Student Policy MoCap}) showed a tremendous improvement in maintaining safe inter-agent distances in many scenarios as shown in Figure \ref{fig:all_drone_dist}. In all real-world deployments, the UAVs never came closer than 0.8 m to each other. 

\subsection{Generalize to similar trajectories}
We test the ability of the student policies to transport the load along similar but different trajectories.
To do this, we train the student policies on two trajectories and then evaluate the resulting policy in simulation on multiple unseen trajectories of \textit{figure-eight}. Training alternated between these two trajectories over 20 DAgger rounds, with perturbations applied during the final 4 rounds. The entire training process for this experiment took 80 minutes.

\begin{figure*}[htbp]
    \centering
    \begin{subfigure}[b]{0.48\textwidth}
        \centering
        \includegraphics[width=\linewidth]{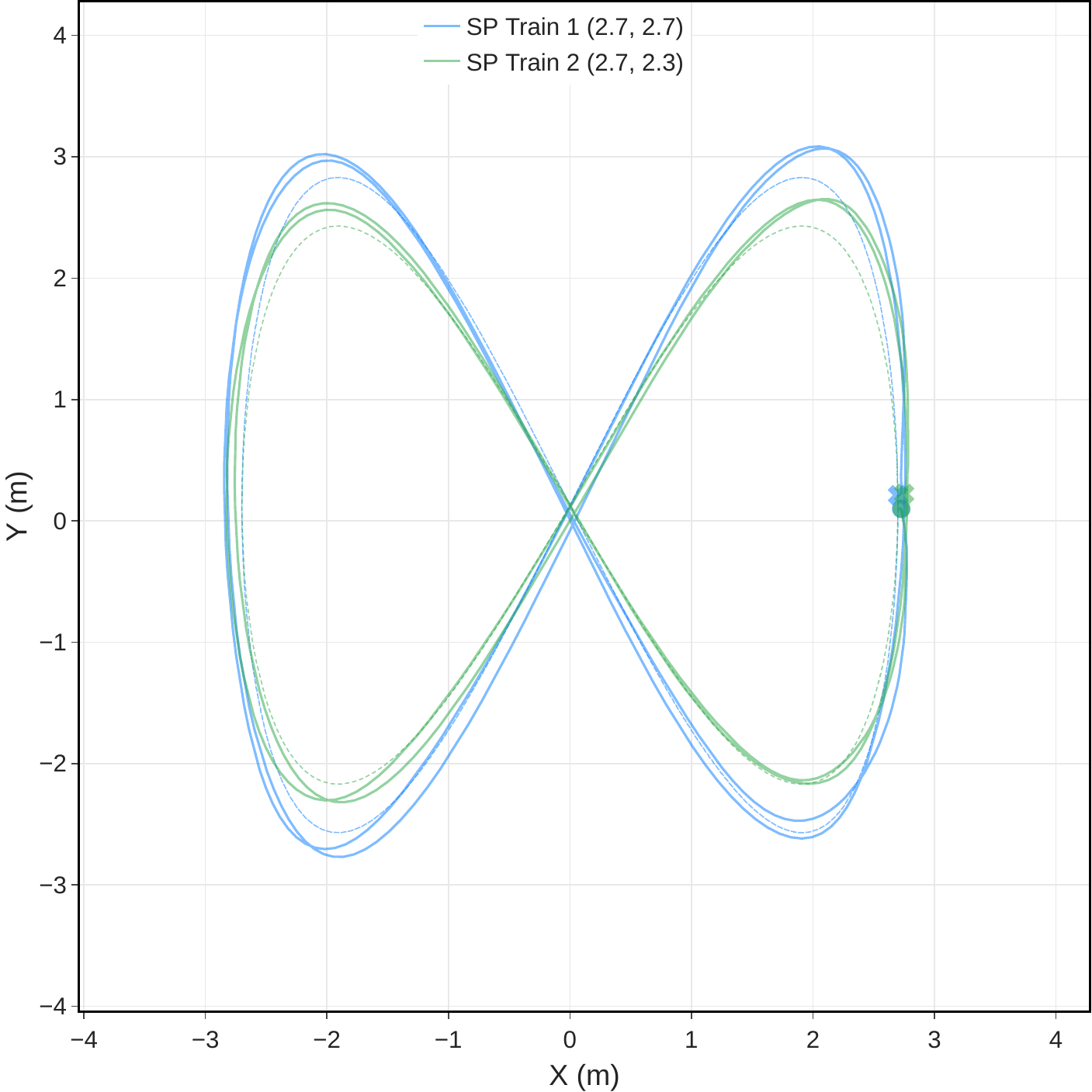}
        \caption{Evaluation of student policies on training trajectories}
    \end{subfigure}
    \hfill
    \begin{subfigure}[b]{0.48\textwidth}
        \centering
        \includegraphics[width=\linewidth]{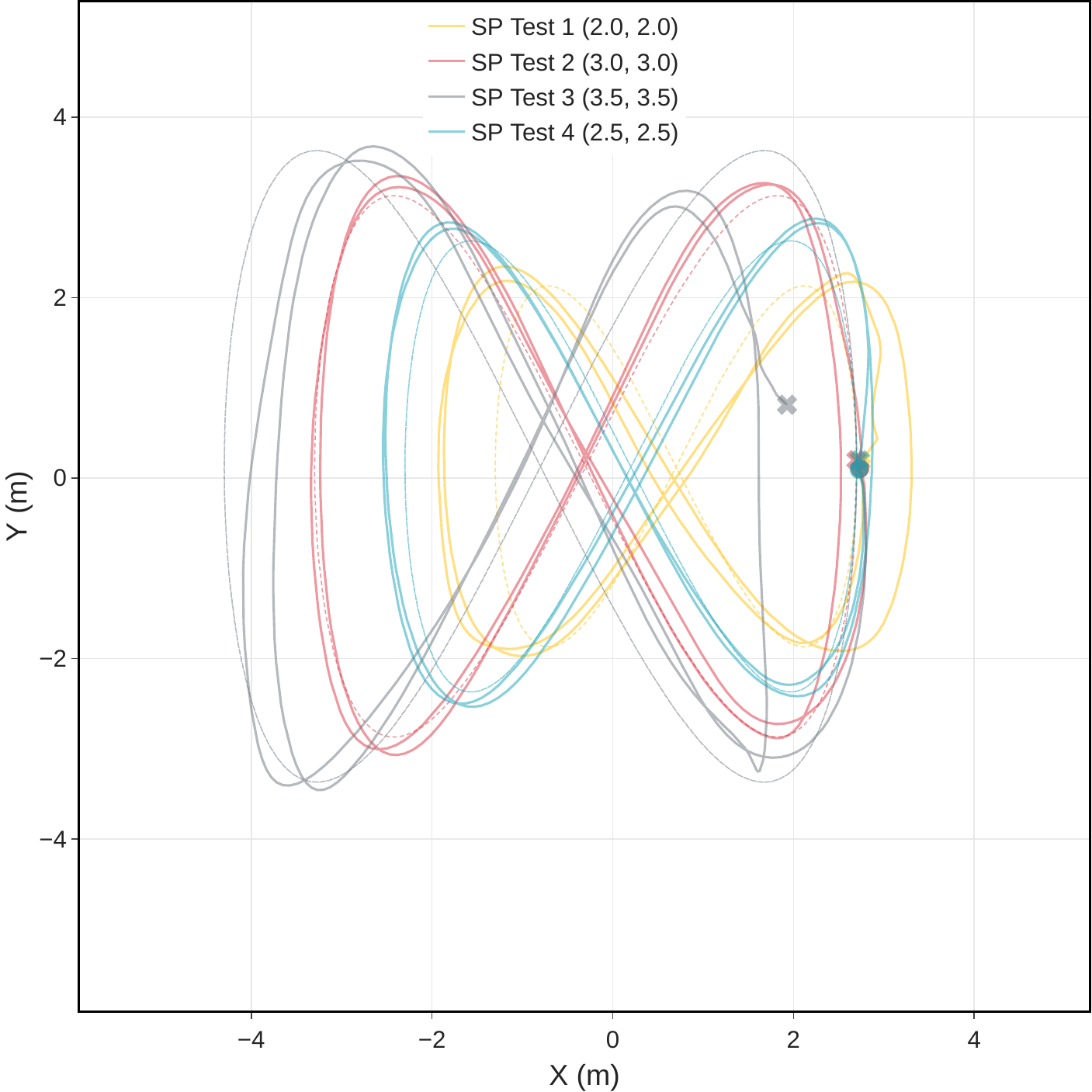}
        \caption{Evaluation of student policies on testing trajectories}
    \end{subfigure}

    \vspace{0.5cm}

    \begin{subfigure}[b]{0.48\textwidth}
        \centering
        \includegraphics[width=\linewidth]{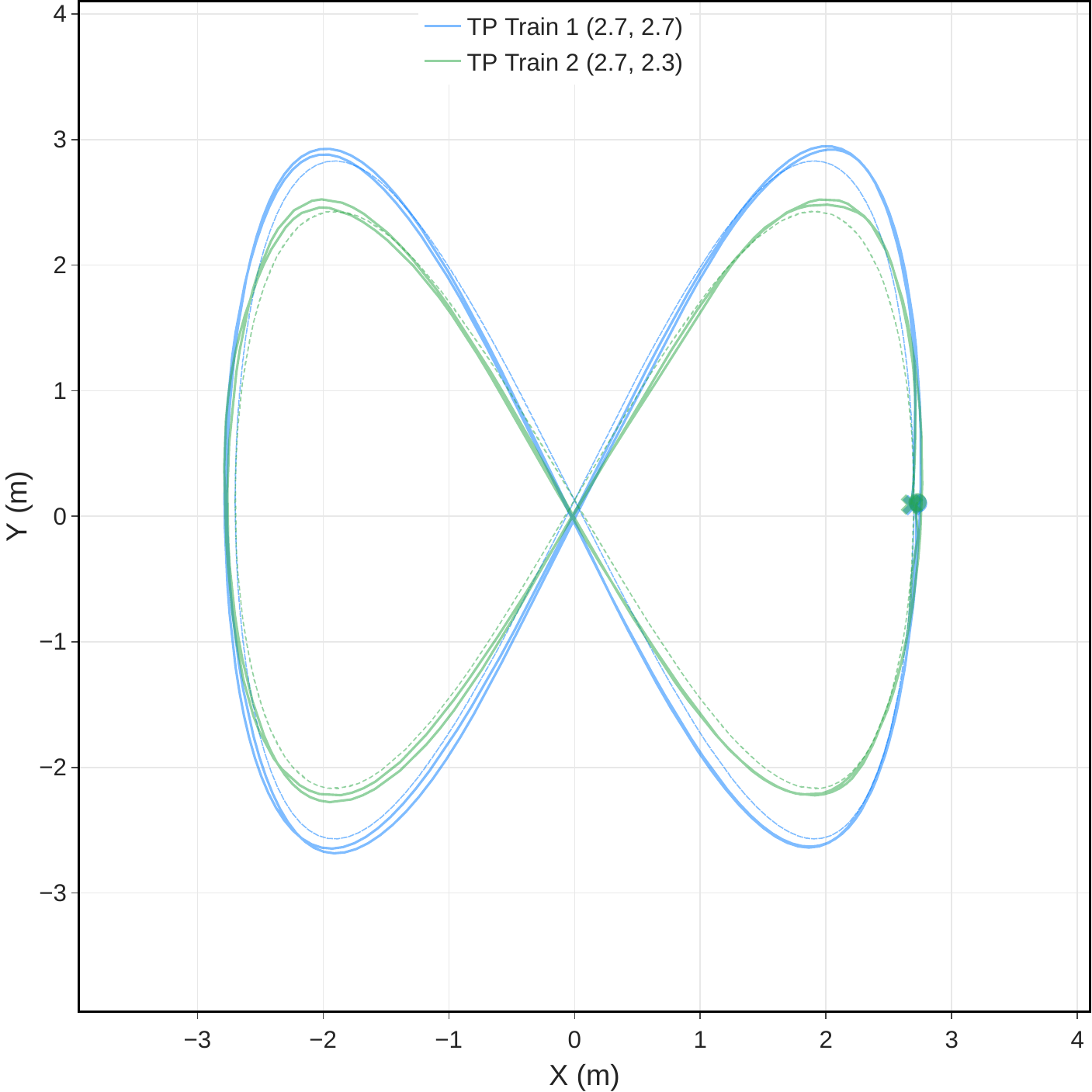}
        \caption{Evaluation of teacher policy on training trajectories}
    \end{subfigure}
    \hfill
    \begin{subfigure}[b]{0.48\textwidth}
        \centering
        \includegraphics[width=\linewidth]{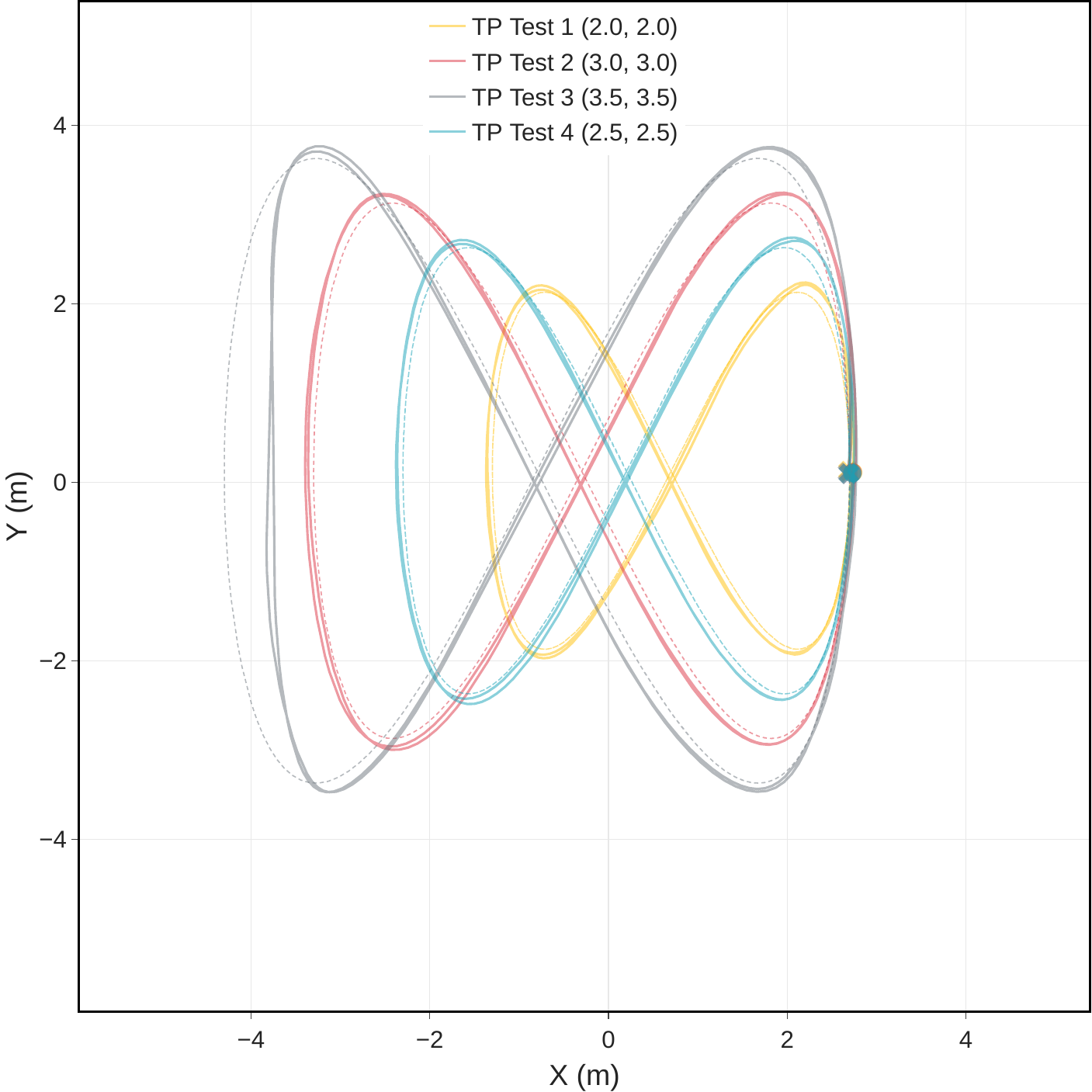}
        \caption{Evaluation of teacher policy on testing trajectories}
    \end{subfigure}
    \caption{Top-down view of tracking performance of the student and teacher policy. Dotted lines represent reference trajectory. Trajectories start and stop at the same point. Same colours represent same trajectories across figures. The performance of the teacher policy on these trajectories shows the ability of the system to execute these trajectories. The student policies struggle to generalize well to unseen trajectories. For the smallest (yellow) trajectory, the student policy struggles to make sharp turns. On the largest (gray) trajectory, the students struggle to maintain the required speeds leading to large pose tracking error.}
    \label{fig:generalizability_topview}
\end{figure*}

The performance of the student policies are summarized in Table \ref{tab:figure8_split2}. The top view visualization of the student is provided in Figure \ref{fig:generalizability_topview}.
We provide the performance of the teacher policy on these trajectories as a baseline.

From this experiment we conclude that student policies show limited ability to generalize to very different trajectories. This may be due to the limited number and kind of trajectories used in training.

\begin{table*}[h]
\centering
\begin{tabularx}{\linewidth}{|X|X|X|cc|cc|}
\cline{1-7}
\multicolumn{2}{|c|}{\textbf{Ref. Size}} & \textbf{Train} & \multicolumn{2}{c|}{\textbf{Student Policies}} & \multicolumn{2}{c|}{\textbf{Teacher Policy}} \\
\cline{1-2} \cline{4-7}
\textbf{$\hat{x_a}$ (m)} & \textbf{$\hat{y_a}$ (m)} & \textbf{Dataset} & Posn. (m) & Orient. ($^\circ$) & Posn. (m) & Orient. ($^\circ$) \\
\cline{1-7}
2.7 & 2.7 & Yes & 0.166 & 6.768  & 0.089 & 4.111 \\
2.7 & 2.3 & Yes & 0.155 & 5.684  & 0.091 & 4.324 \\
2.0 & 2.0 & No  & 0.377 & 10.877 & 0.073 & 3.397 \\
3.0 & 3.0 & No  & 0.175 & 7.531  & 0.105 & 5.371 \\
3.5 & 3.5 & No  & 0.707 & 12.013 & 0.192 & 6.912 \\
2.5 & 2.5 & No  & 0.205 & 6.903  & 0.090 & 4.196 \\
\cline{1-7}
\end{tabularx}
\caption{Comparison of the student and teacher policy performance in simulation for multiple trajectories. The table reports the RMSE for position and orientation. Student policies were trained using the first two trajectories. $\hat{x_a}$ \& $\hat{y_a}$ denote the amplitudes of the sine and cosine functions used to generate the trajectories.}
\label{tab:figure8_split2}
\end{table*}

\begin{figure}
    \centering
    \begin{subfigure}[b]{0.48\textwidth}
        \centering
        \includegraphics[width=\linewidth]{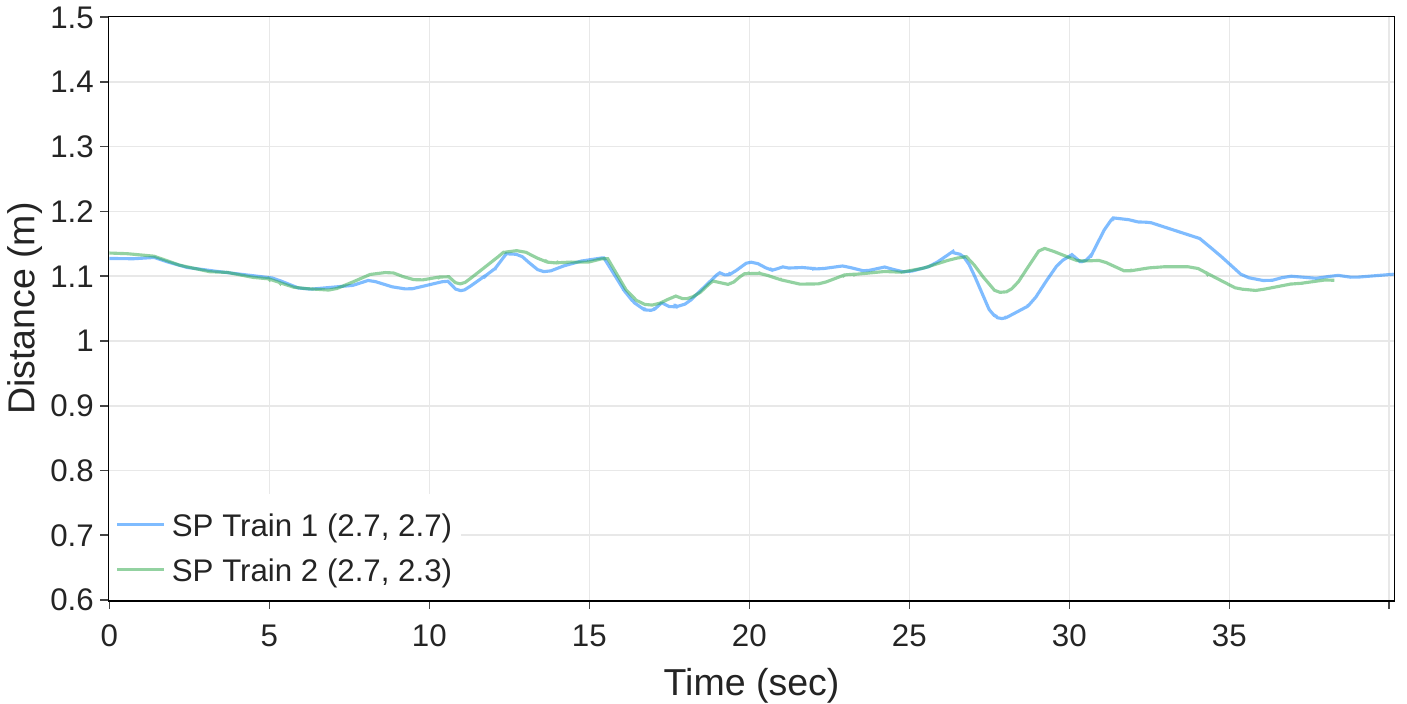}
    \end{subfigure}
    \hfill
    \begin{subfigure}[b]{0.48\textwidth}
        \centering
        \includegraphics[width=\linewidth]{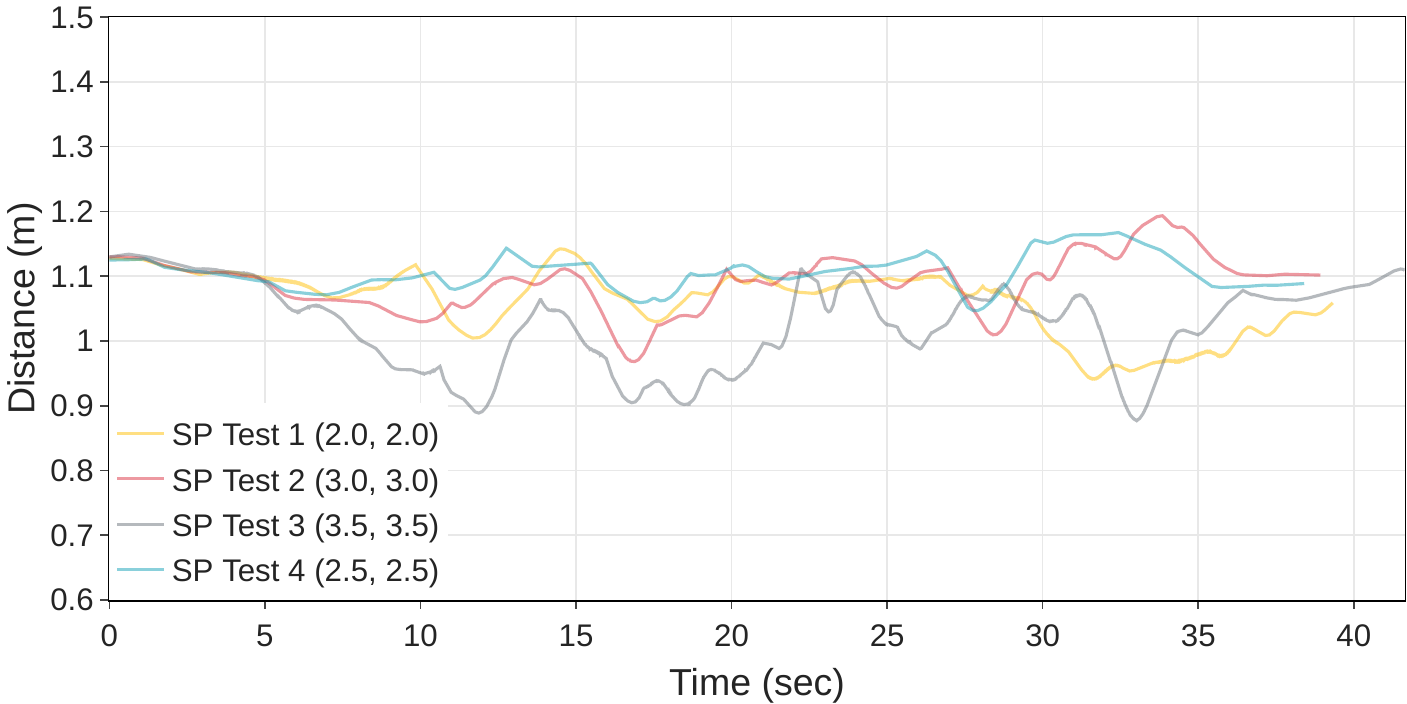}
    \end{subfigure}
    \caption{Inter-agent distance are shown here. The quadrotors are able to maintain safe inter-agent distances in all tested scenarios despite lacking awareness of the position of other UAVs.}
    \label{fig:generalize_1_drone_dist}
\end{figure}

\subsection{Generalize to dissimilar trajectories}
We also train the student policy to learn very different kinds of trajectories—namely, Eight, Circle, and Square. During training, we randomly sample one of these trajectory types at each iteration to encourage the student policy to generalize across diverse motion patterns. These set of trajectories contain a very diverse set of curves with a large range of radii of curvature which needs to be followed by the load. The z value of the trajectories also oscillates with a random amplitude and frequency. The trajectories can also have multiple modes for the desired orientation. These modes further increases the complexity of the task significantly.

\subsubsection{Constant orientation}
Trajectories with constant orientation of the load result in UAV trajectories that are very similar to the given trajectory. The student policies need to focus only on position tracking which leads to lower errors.

\subsubsection{Zero-sideslip (ZSS) orientation}
\begin{figure*}[htbp]
    \centering
    \begingroup
    \captionsetup[subfigure]{font=tiny}
    \begin{subfigure}[b]{0.36\textwidth}
        \centering
        \includegraphics[width=\linewidth]{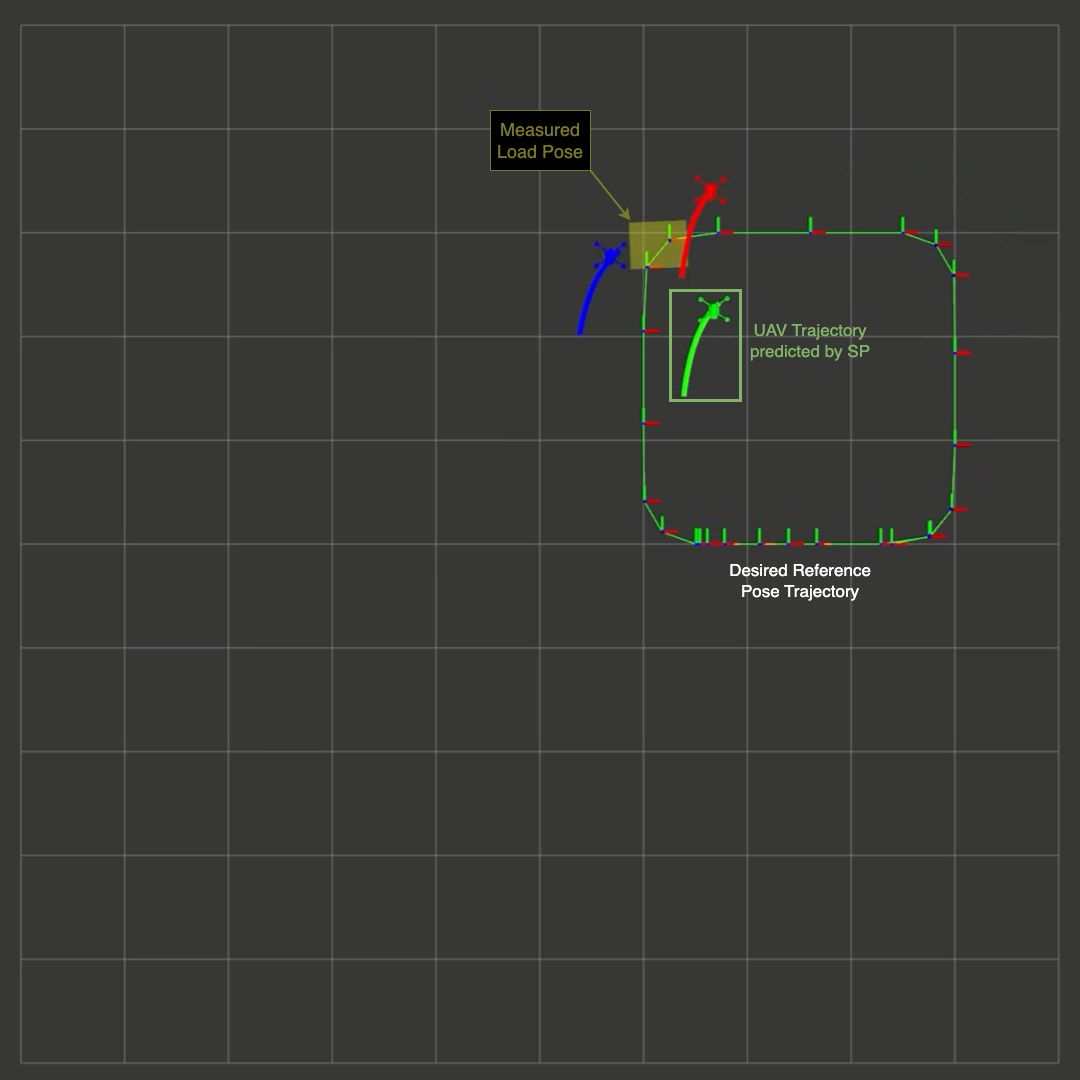}
        \caption{Square constant-orientation trajectory: Student policy for each UAV independently predicts the drone path for the ego UAV. This path is tracked using the onboard low-level controllers on the UAV.}
    \end{subfigure}
    \hfill
    \begin{subfigure}[b]{0.36\textwidth}
        \centering
        \includegraphics[width=\linewidth]{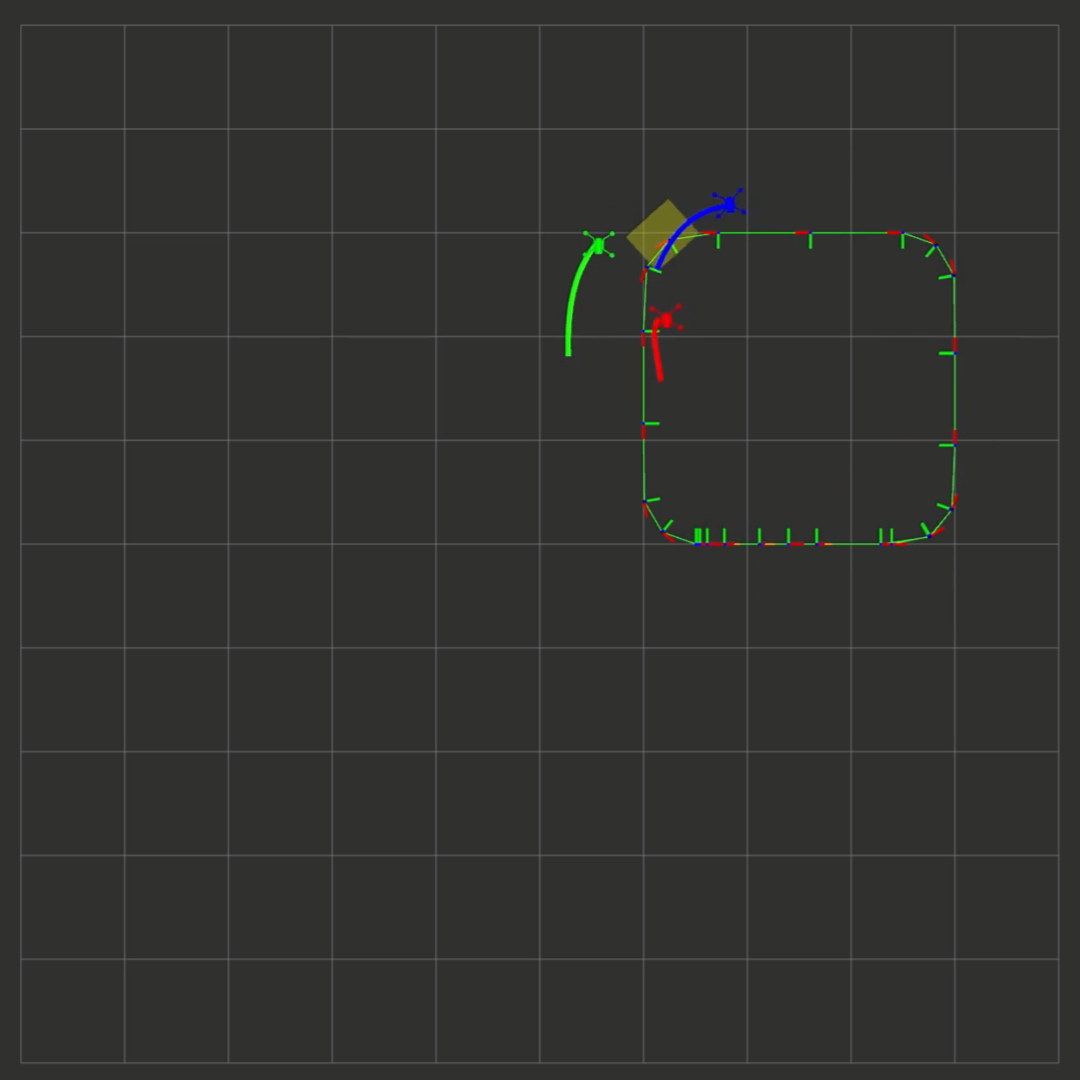}
        \caption{Square zero-sideslip trajectory: The desired body X axis is aligned along velocity. Notice how the inner UAV needs to slow down while the outer UAVs accelerate to track the desired pose.}
    \end{subfigure}

    \begin{subfigure}[b]{0.36\textwidth}
        \centering
        \includegraphics[width=\linewidth]{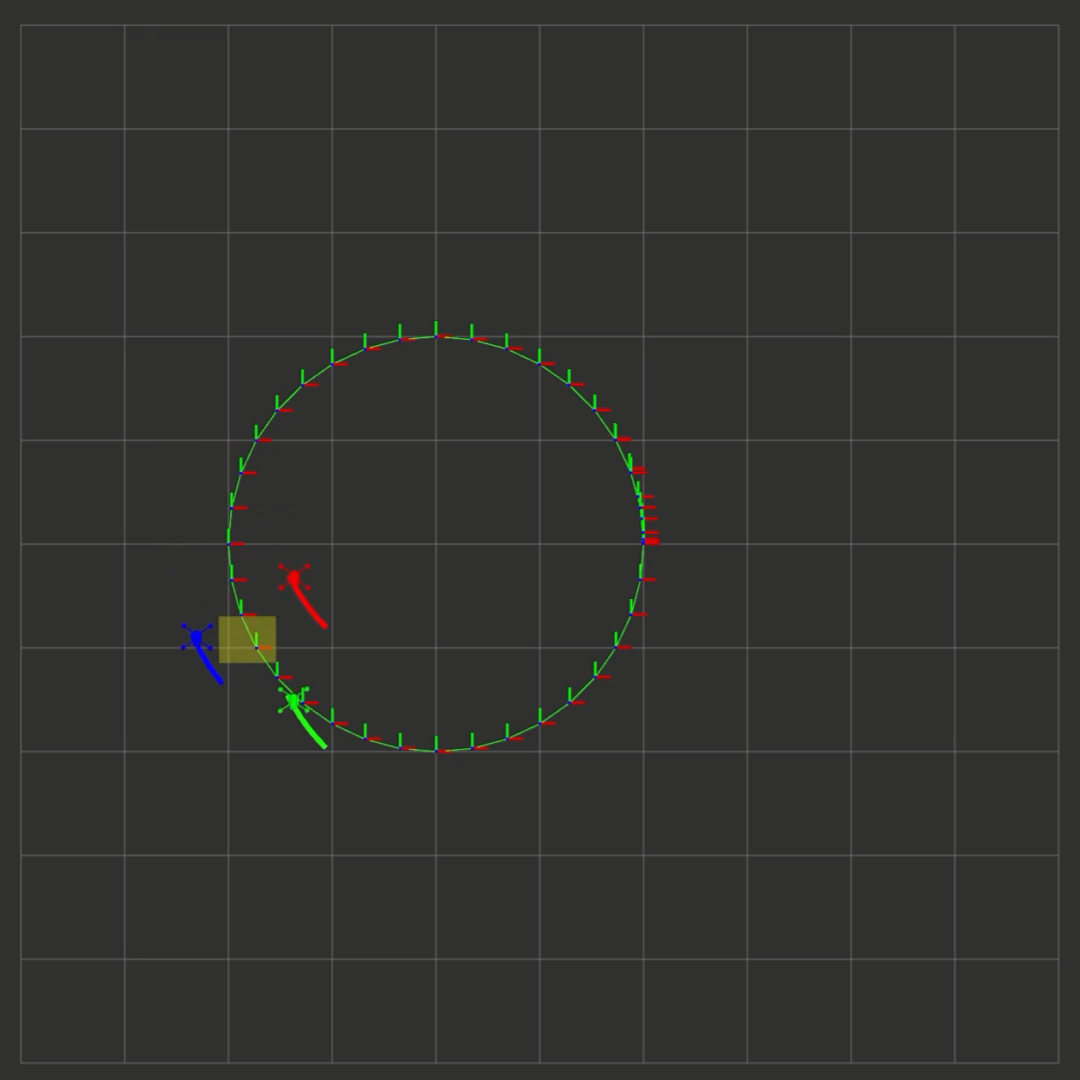}
        \caption{Circle constant orientation: UAVs produce similar trajectories as the orientation of the load is constant.}
        \label{fig:circle_no_yaw}
    \end{subfigure}
    \hfill
    \begin{subfigure}[b]{0.36\textwidth}
        \centering
        \includegraphics[width=\linewidth]{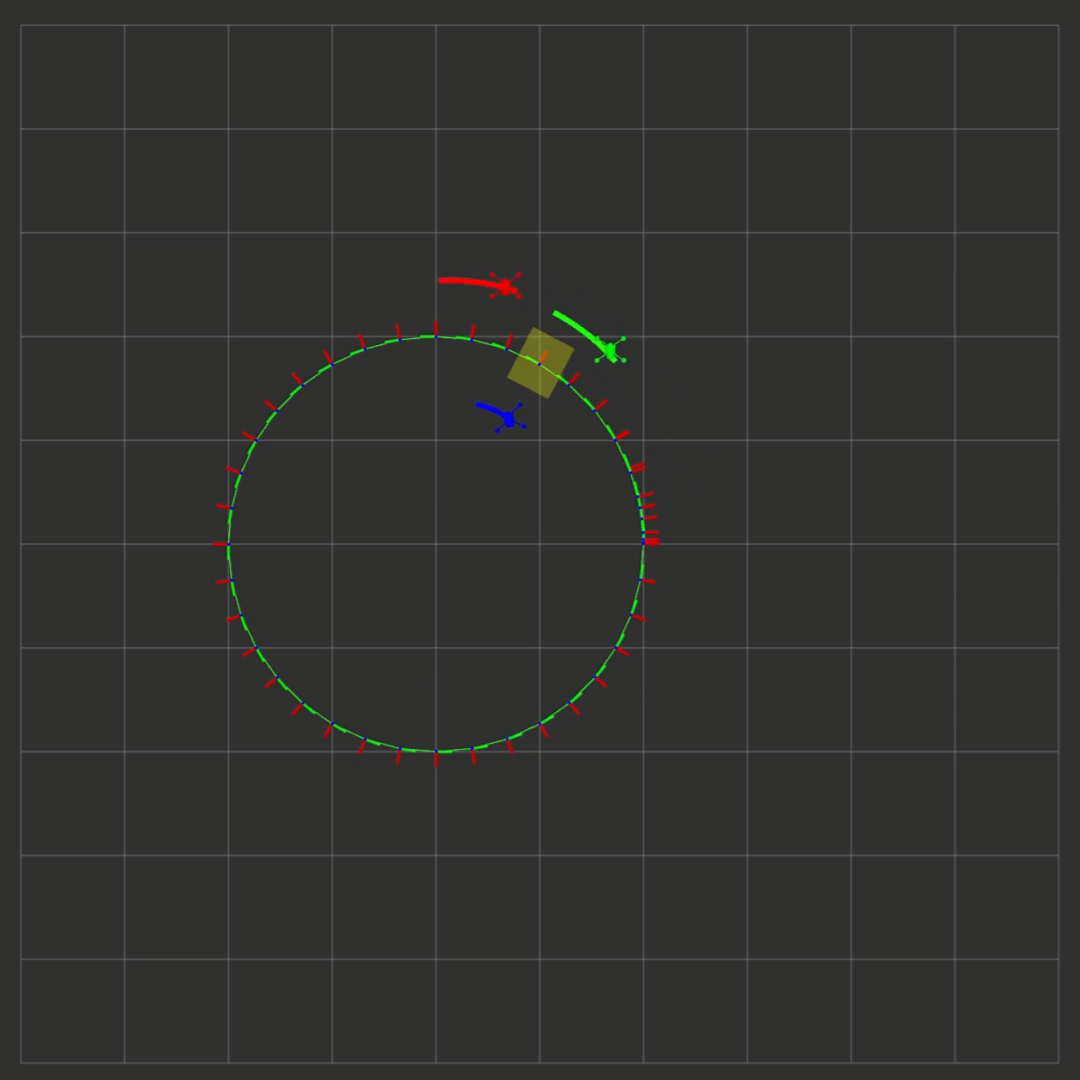}
        \caption{Circle zero-sideslip trajectory: The formation of the UAV is very different to Figure \ref{fig:circle_no_yaw} due to the desired orientation. }
    \end{subfigure}

    \begin{subfigure}[b]{0.36\textwidth}
        \centering
        \includegraphics[width=\linewidth]{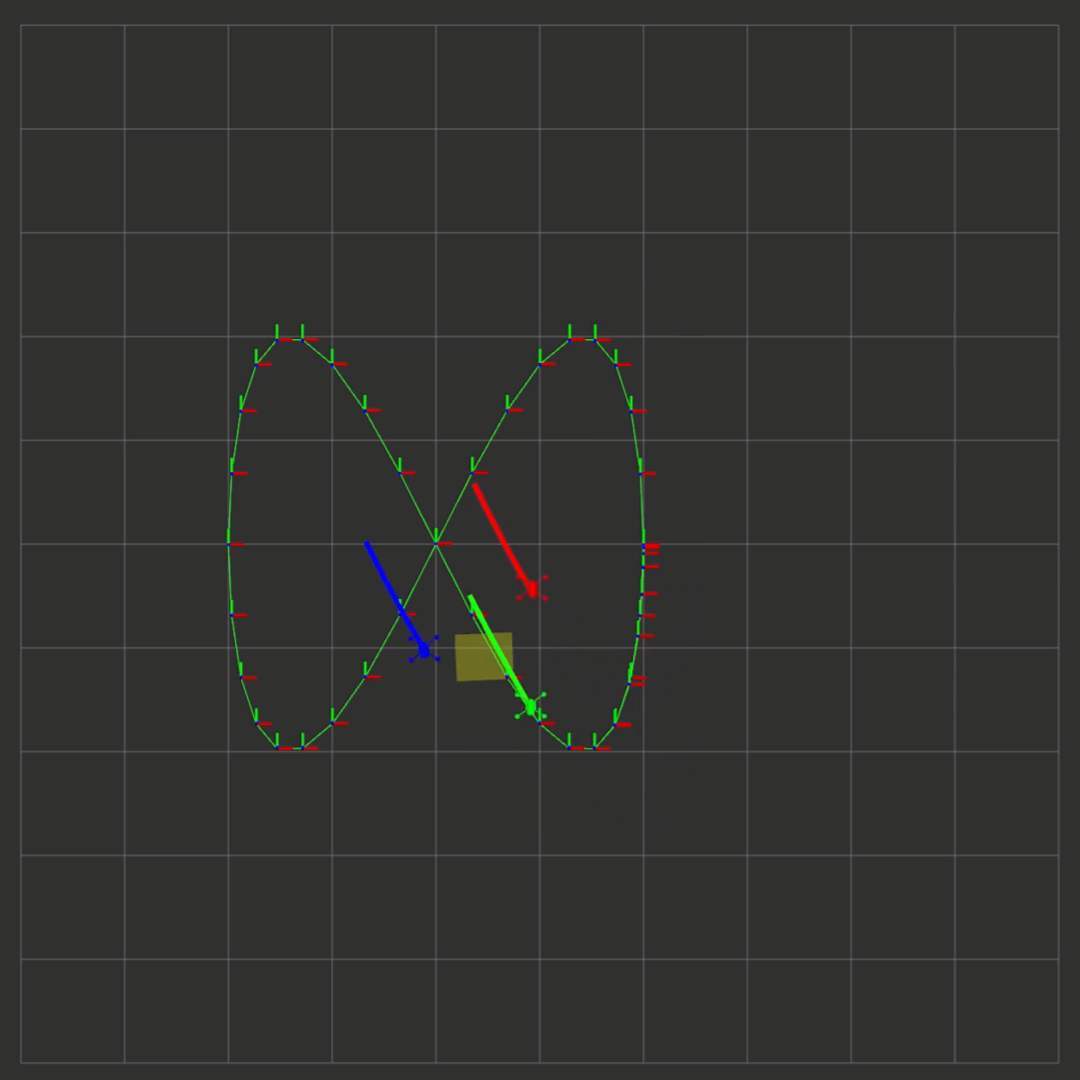}
        \caption{Eight constant orientation: UAVs produce similar trajectories as the orientation of the load is constant. \\[5.2ex]}
        \label{fig:eight_no_yaw}
    \end{subfigure}
    \hfill
    \begin{subfigure}[b]{0.36\textwidth}
        \centering
        \includegraphics[width=\linewidth]{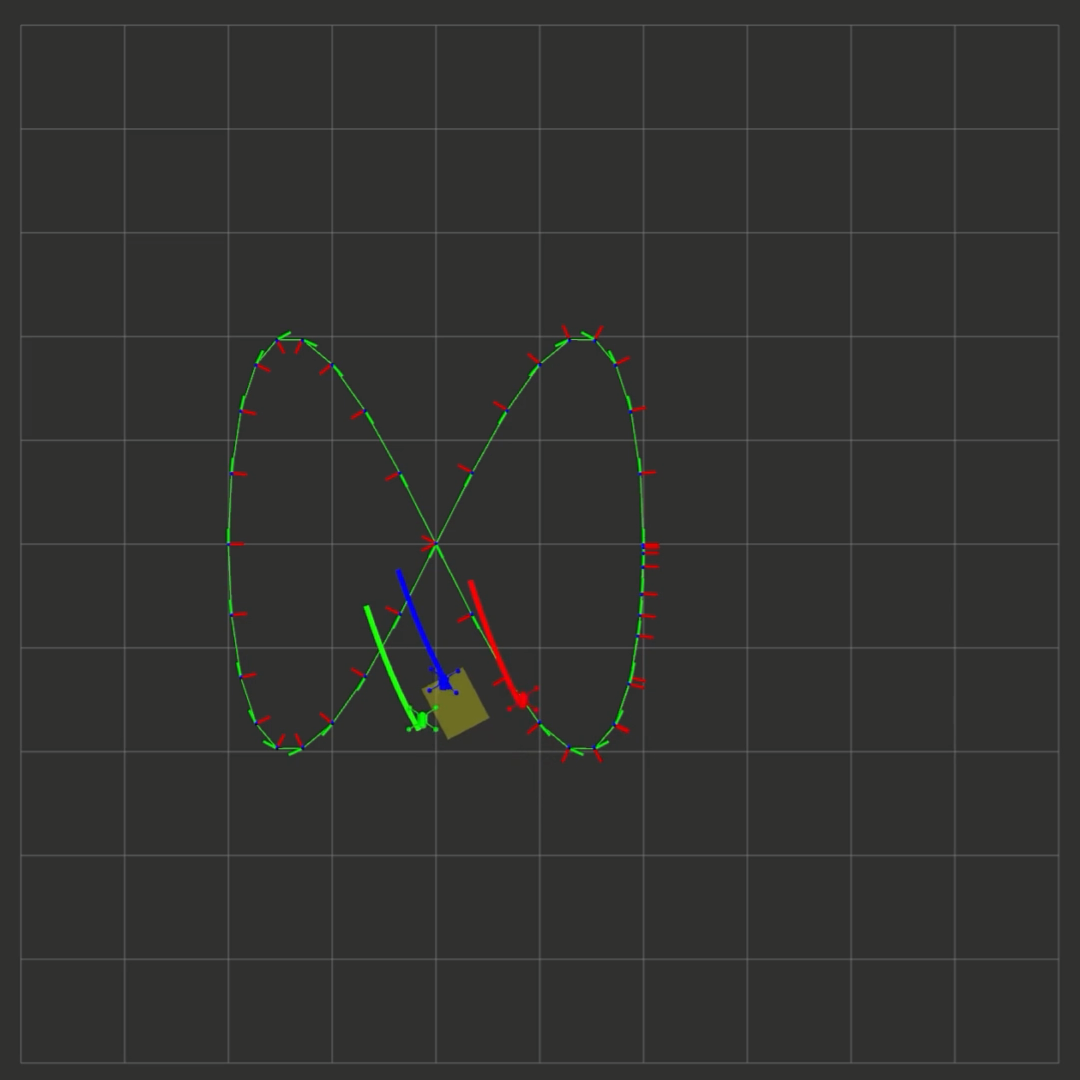}
        \caption{Eight zero-sideslip trajectory: The policy struggles to adapt to the rapid changes in the payload's desired orientation. As a result, the blue and green drones converge too closely, prompting the safety module to intervene.}
        \label{fig:eight_yaw}
    \end{subfigure}
    \endgroup

    \caption{Visualization of the performance Student Policy (Both) on various trajectories. This policy is capable of predicting vastly different UAV trajectory based on the desired reference trajectory.}
    \label{fig:tight_turns}
\end{figure*}

To maintain the correct orientation for these trajectories, all the UAVs need to follow very different trajectories. For executing tight turns, the inner UAV needs to slow down or even reverse while the UAV on the outside needs to move fast along an arc. The figure \ref{fig:tight_turns} shows some examples. 

\begin{table*}[h]
    \centering
    \scriptsize
    \begin{tabular}{|l|l|cc|cc|cc|}
    \hline
    \textbf{Trajectory} & \textbf{Ref Size (m)} & \multicolumn{2}{c|}{\textbf{SP (Constant)}} & \multicolumn{2}{c|}{\textbf{SP (Zero-sideslip)}} & \multicolumn{2}{c|}{\textbf{SP (Both)}} \\
     &  & Posn (m) & Orient. ($^\circ$) & Posn (m) & Orient. ($^\circ$) & Posn (m) & Orient. ($^\circ$) \\
    \hline
    Eight (Constant) & $\hat{x_a}$: 2, $\hat{y_a}$:: 2 & 0.097 & 4.443 & \textcolor{red}{\ding{55}} & \textcolor{red}{\ding{55}} & 0.089 & 4.947 \\
    Eight (Zero-sideslip) & $\hat{x_a}$: 2, $\hat{y_a}$:: 2 & \textcolor{red}{\ding{55}} & \textcolor{red}{\ding{55}} & 0.212 & 12.603 & \textcolor{red}{\ding{55}} & \textcolor{red}{\ding{55}} \\
    Circle (Constant) & Radius: 2 & 0.037 & 2.850 & \textcolor{red}{\ding{55}} & \textcolor{red}{\ding{55}} & 0.049 & 3.518 \\
    Circle (Zero-sideslip) & Radius: 2 & \textcolor{red}{\ding{55}} & \textcolor{red}{\ding{55}} & 0.095 & 8.293 & 0.134 & 10.403 \\
    Square (Constant) & Side: 2 & 0.202 & 5.475 & \textcolor{red}{\ding{55}} & \textcolor{red}{\ding{55}} & 0.149 & 6.540 \\
    Square (Zero-sideslip) & Side: 2 & \textcolor{red}{\ding{55}} & \textcolor{red}{\ding{55}} & 0.169 & 14.803 & 0.181 & 13.107 \\
    \hline
    \end{tabular}
    \caption{Comparison of performance of various Student Policies on Different Trajectories. \textcolor{red}{\ding{55}} represents instances where inter-agent distance reduced to less than 0.4 m after which the experiment was stopped.}
    \label{tab:student_policy_comparison}
\end{table*}

We trained three student policies on these trajectories:
\begin{itemize}
    \item \textbf{Student Policy (Constant)}: Trained on all constant orientation trajectories for 30 rounds. This policy is unable to follow trajectories where the orientation changes.
    \item \textbf{Student Policy (Zero-sideslip)}: Trained on all zero-sideslip trajectories for 30 rounds. This policy is unable to follow trajectories where the orientation is constant.
    \item \textbf{Student Policy (Both)}: Trained on all trajectories with zero-sideslip and constant orientation for 60 rounds. This unified policy can track all trajectories except the figure-eight zero-sideslip path, though it exhibits higher pose tracking error compared to specialized policies.
\end{itemize}

\begin{figure}[h]
    \centering
    \begin{subfigure}[b]{0.48\textwidth}
        \centering
        \includegraphics[width=\linewidth]{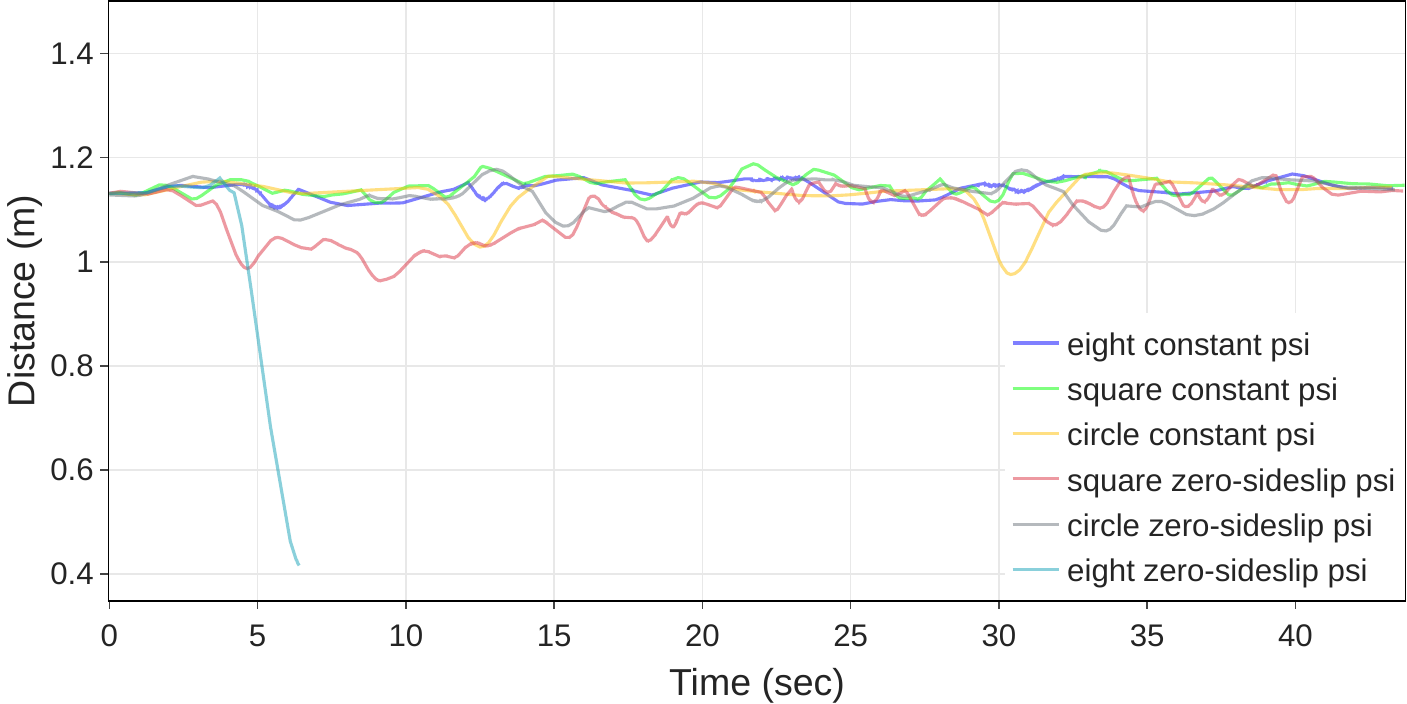}
        \caption{Student Policy (Both): Inter-agent distances for various constant orientation and zero-sideslip trajectories. The policy maintains safe inter-agent distance for all trajectories except the figure-eight zero-sideslip trajectory.}
        \label{fig:pinn_v2_generalize2_drone_dist}
    \end{subfigure}
    \hfill
    \begin{subfigure}[b]{0.48\textwidth}
        \centering
        \includegraphics[width=\linewidth]{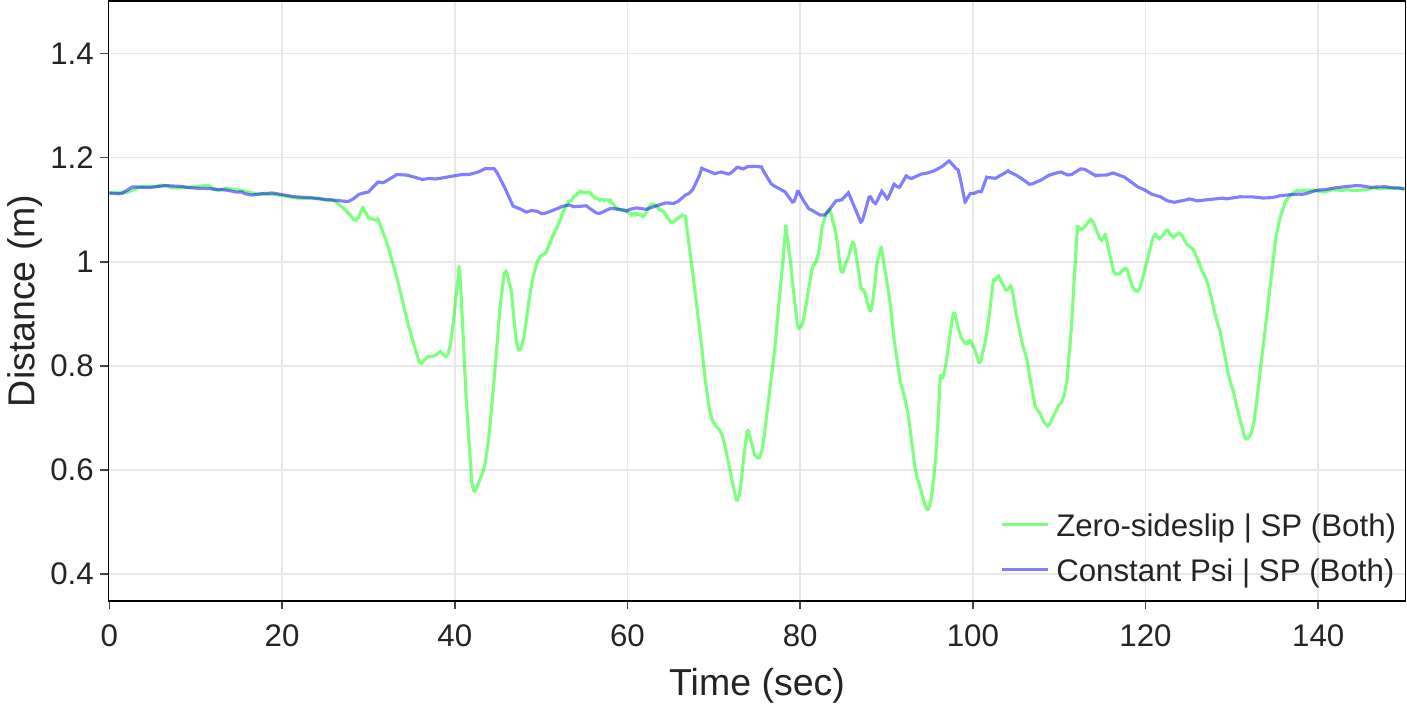}
        \caption{Student Policy (Both): The inter-agent distances on the Zandvoort F1 track are illustrated here. During zero-sideslip trajectories, the agents frequently come close to one another but consistently maintain a distance of more than 0.4 meters.}
        \label{fig:zandvoort_drone_dist}
    \end{subfigure}
    \caption{We show inter-agent distance graphs for the Student Policy (Both) on various trajectories.}
    \label{fig:generalize_2_drone_dist}
\end{figure}

The performance of these models is summarised in Table \ref{tab:student_policy_comparison}. The performance of the Student Policy (Both) is also visualised in Fig. \ref{fig:tight_turns} and Fig. \ref{fig:pinn_v2_generalize2_drone_dist}.

\subsection{Zandvoort F1 Track}
The proposed student policy was evaluated on its capability to transport the payload along the Formula 1 (F1) track located in Zandvoort, Netherlands.
This trajectory differs substantially from those encountered during training, exhibiting greater variability in curvature radii, path length, and overall duration.

\begin{figure}[h!]
    \centering
    \begin{subfigure}[b]{0.48\textwidth}
        \centering
        \includegraphics[width=\linewidth]{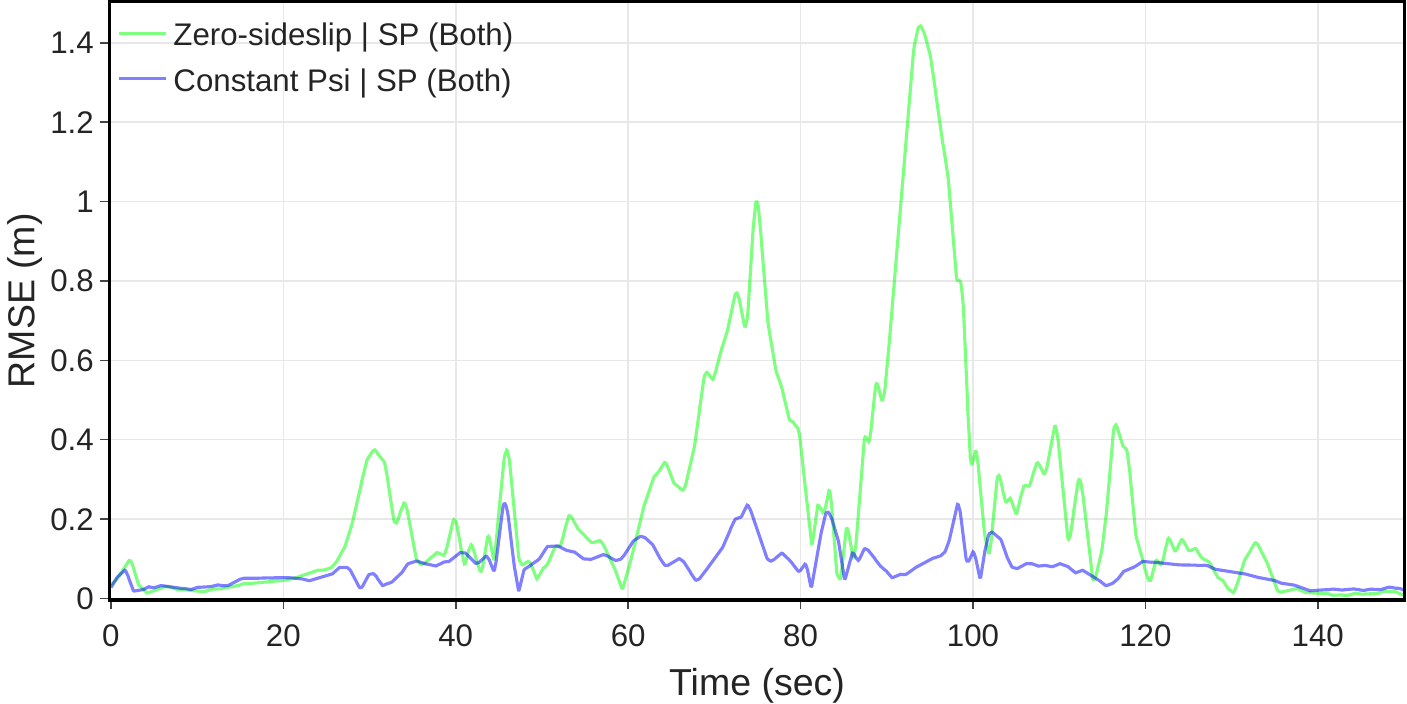}
        \caption{Position tracking error signal}
        \label{fig:zandvoort_att}
    \end{subfigure}
    \hfill
    \begin{subfigure}[b]{0.48\textwidth}
        \centering
        \includegraphics[width=\linewidth]{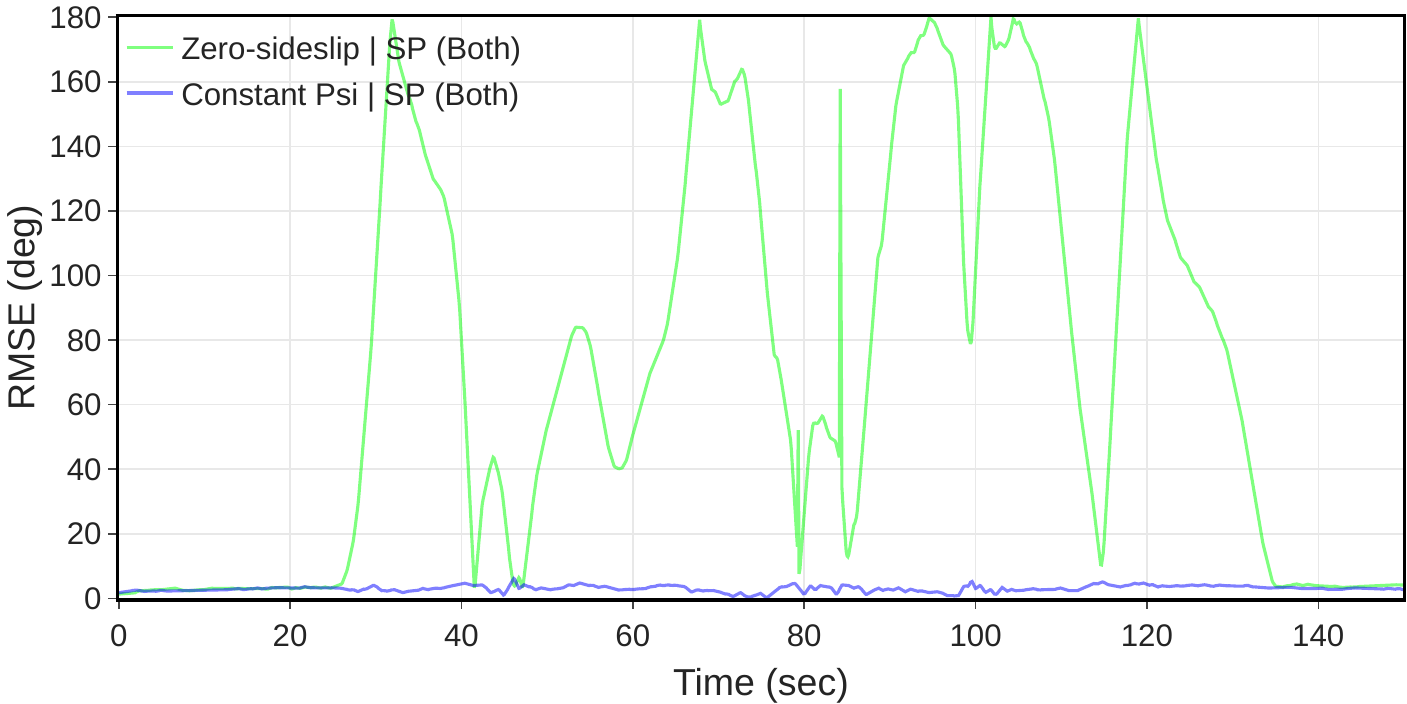}
        \caption{Attitude tracking error signal}
        \label{fig:zandvoort_posn}
    \end{subfigure}
    \caption{Time-series of error signals when tracking the Zandvoort F1 Track using the Student Policy (Both).}
    \label{fig:zandvoort_posn_att}
\end{figure}

\begin{figure}[h!]
    \centering
    \includegraphics[width=0.8\linewidth]{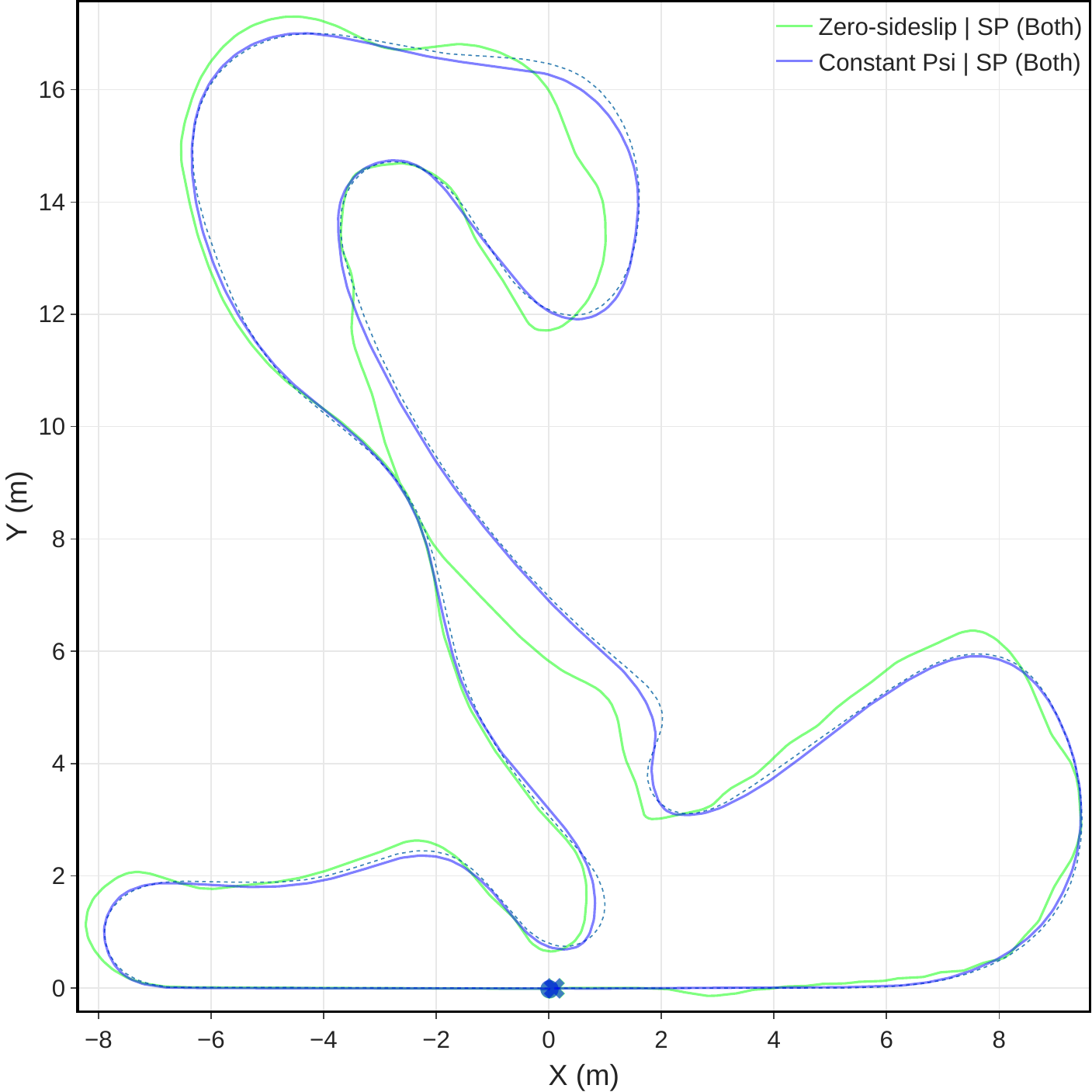}
    \caption{Top-down view of tracking performance for the Zandvoort F1 Track using the Student Policy (Both).}
    \label{fig:zandvoort_load_ref_tracking}
\end{figure}

We evaluated the Student Policy (SP (Both)) under both orientation modes: constant orientation and zero-sideslip. The quantitative results are summarized in Table \ref{tab:zandvoort_perf} and illustrated in Figure \ref{fig:zandvoort_posn_att}, \ref{fig:zandvoort_load_ref_tracking}.

\begin{table}[h]
    \centering
    \begin{tabular}{|c|c|c|}
    \cline{1-3}
    \textbf{Metric}        & \textbf{Zandvoort (Constant)} & \textbf{Zandvoort (ZSS)} \\
    \cline{1-3}
    RMSE Posn. (m)   & 0.092      & 0.378   \\
    RMSE Orient. ($^\circ$)  & 3.157   & 93.575 \\
    \cline{1-3}
    \end{tabular}
    \caption{Comparison of metrics from simulated flights around the Zandvoort F1 Track. The Student Policy (Both) was deployed for this experiment without any fine-tuning.}
    \label{tab:zandvoort_perf}
\end{table}

In the constant orientation mode, the proposed method achieved high position-tracking accuracy, demonstrating the policy’s ability to compute appropriate ego-UAV positions for diverse trajectories. 
The policies exhibit invariance to trajectory scale and duration, as the student processes fixed 2-second trajectory segments at each time step.
Furthermore, by expressing observations in the load frame $\mathcal{L}$, the policy is equivariant to the absolute position of the system in the world allowing it to operate in arbitrarily large spaces.

In contrast, in the zero-sideslip mode, the student policy exhibited significantly higher position-tracking error and failed to accurately follow the desired orientation. This suggests that the policy struggles to determine the appropriate UAV positions for novel trajectories involving varying orientations. This is likely due to overfitting to the limited set of three trajectory types used for training as discussed in the preceding section.

\subsection{Full Pose Control}
This section evaluates the student policy's capability to manipulate the complete pose of a payload. The policy receives a desired goal pose represented as a 6-dimensional vector (position and Euler angles) as input. A smooth trajectory is generated between the current payload pose (starting point) and the desired goal pose (endpoint), ensuring zero velocities at both trajectory endpoints.
For this experiment, we trained a student policy over 80 DAgger iterations. Each iteration comprised a single episode in which the payload was manipulated from its current starting pose to a randomly selected target pose. The starting pose for each iteration corresponded to the final payload pose achieved in the preceding experiment. The target poses were uniformly sampled within $\pm15^{o}$ for roll and pitch angles, with yaw angles sampled across the full range. Consequently, the student policy learned to transport the payload across a diverse set of initial and final pose configurations. All trajectories were executed over a 11-second duration.

\subsubsection{Set point Tracking}
\begin{figure*}[h]
    \centering
    \begin{subfigure}[b]{0.32\textwidth}
        \centering
        \includegraphics[width=\linewidth]{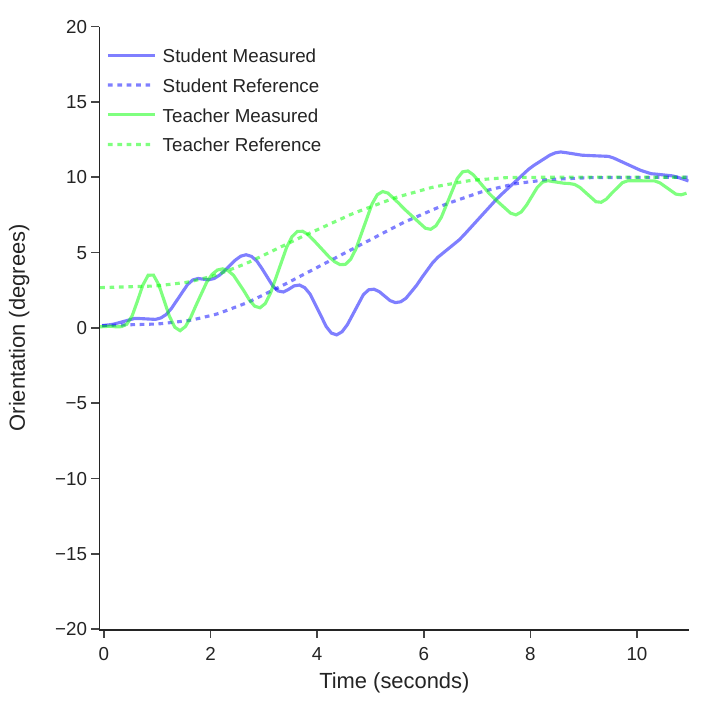}
        \caption{Roll}
    \end{subfigure}
    \begin{subfigure}[b]{0.32\textwidth}
        \centering
        \includegraphics[width=\linewidth]{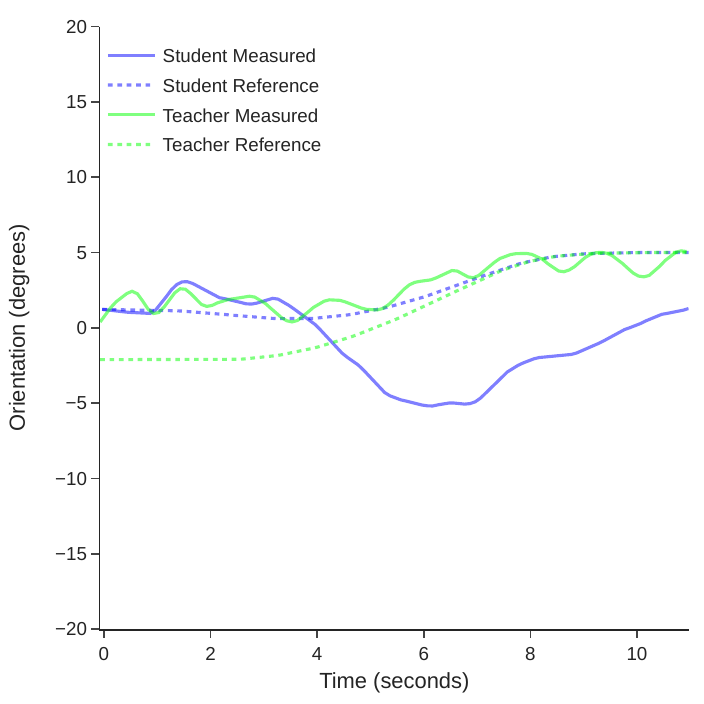}
        \caption{Pitch}
    \end{subfigure}
    \begin{subfigure}[b]{0.32\textwidth}
        \centering
        \includegraphics[width=\linewidth]{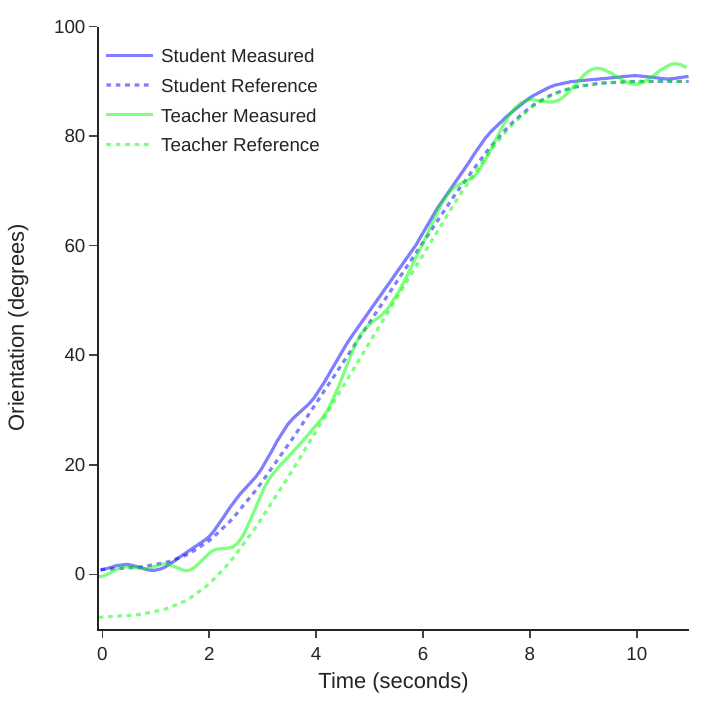}
        \caption{Yaw}
    \end{subfigure}

    \vspace{0.3cm}

    \begin{subfigure}[b]{0.32\textwidth}
        \centering
        \includegraphics[width=\linewidth]{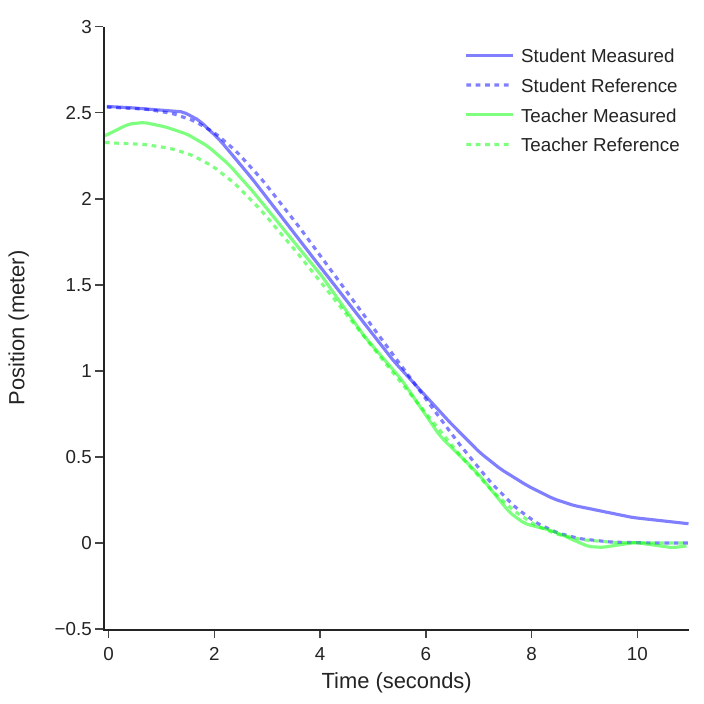}
        \caption{Position X}
    \end{subfigure}
    \begin{subfigure}[b]{0.32\textwidth}
        \centering
        \includegraphics[width=\linewidth]{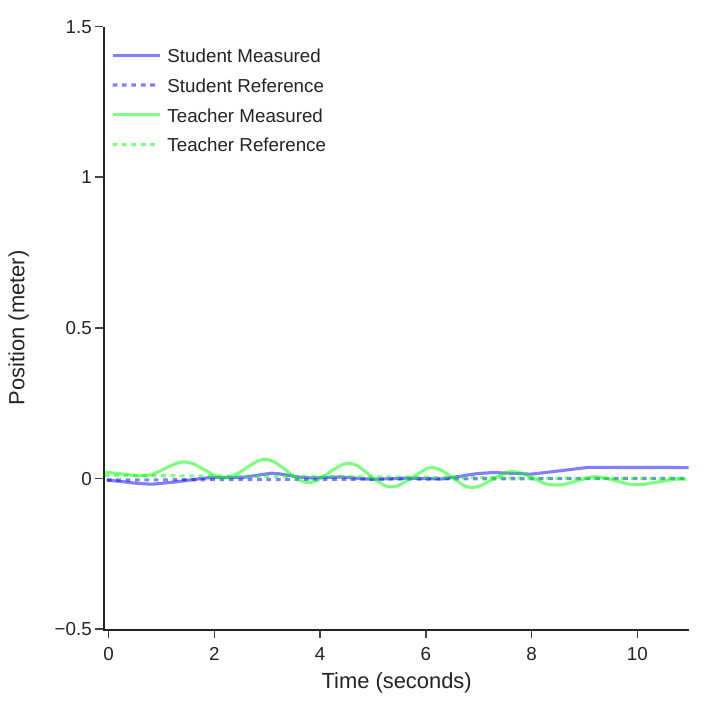}
        \caption{Position Y}
    \end{subfigure}
    \begin{subfigure}[b]{0.32\textwidth}
        \centering
        \includegraphics[width=\linewidth]{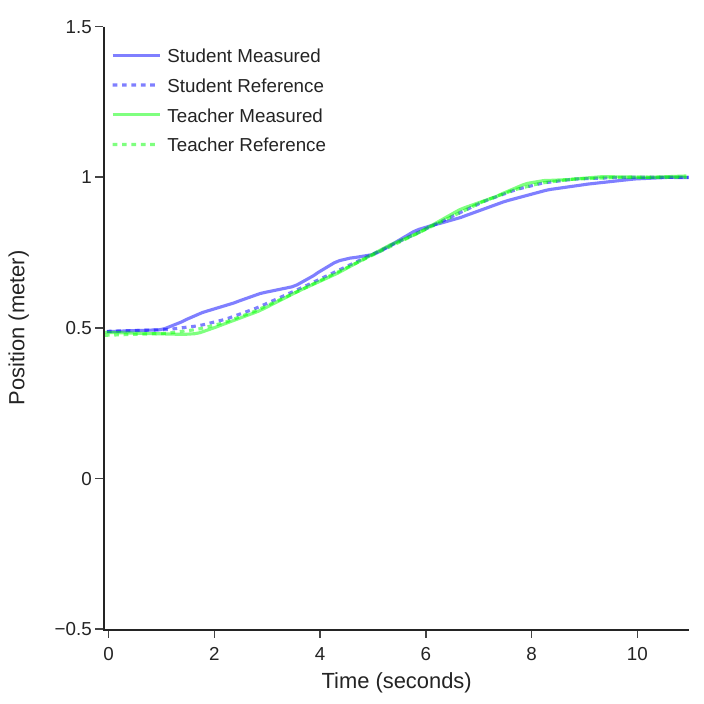}
        \caption{Position Z}
    \end{subfigure}
    \caption{Full pose tracking performance: Orientation (top row) and Position (bottom row) components for the payload along the reference line trajectory. All plots have different Y axis scales. Note that the initial poses differ slightly between the student and teacher policies, as the teacher policy utilizes Extended Kalman Filter (EKF) measurements to establish the starting pose for trajectory generation.}
    \label{fig:full_pose_tracking}
\end{figure*}

We compare the student policy's payload manipulation capabilities against the teacher policy, as illustrated in Figure \ref{fig:full_pose_tracking}. Our analysis demonstrates that the student policy successfully manipulates the payload along the generated reference trajectory. Notably, the student policy achieves smoother payload manipulation compared to the teacher policy, which exhibits oscillatory behaviour due to noisy Extended Kalman Filter (EKF) measurements. However, the centralized teacher policy achieves superior accuracy in the final measured pose. The quantitative results are summarised in Table \ref{tab:pose_tracking_perf}

\begin{table}[h]
    \centering
    \begin{tabular}{|c|c|c|}
    \cline{1-3}
    \textbf{Metric}        & \textbf{Teacher Policy} & \textbf{Student Policies} \\
    \cline{1-3}
    RMSE Posn. (m)   & 0.015      & 0.097   \\
    RMSE Orient. ($^\circ$)  & 1.631   & 3.516 \\
    \cline{1-3}
    \end{tabular}
    \caption{Comparison of pose tracking performance of the Teacher and Student policies.}
    \label{tab:pose_tracking_perf}
\end{table}

\subsubsection{Generalization to Sequential Pose Commands}
This experiment evaluates the student policy's ability to navigate through a sequence of randomly genrated poses. Consistent with the training protocol, the initial pose for each subsequent target depends on the previously achieved pose. Consequently, the starting orientation of the payload is not constrained to be horizontal (i.e., zero pitch and roll angles). Table \ref{tab:generalize_poses} presents the performance metrics across multiple experimental trials. The results demonstrate an average position error of 0.058 meters and an average orientation error of 3.590 degrees.

\begin{table*}[ht]
\centering
\scriptsize
\renewcommand{\arraystretch}{1.3}
\setlength{\tabcolsep}{6pt}
\begin{tabular}{|c|c|c|c|c|c|}
\hline
\textbf{S. No} & \textbf{Type} & \textbf{Start Pose} & \textbf{Goal Pose} & \textbf{Measured End Pose} & \textbf{Error} \\
\hline
\multirow{2}{*}{0} 
& Position:
& 2.5465, 0.0034, 0.4888
& -0.1662, -1.4837, 1.2471
& -0.1520, -1.2842, 1.2322
& 0.2005 \\
& Orientation: 
& 0.7105, 1.1803, 0.3724
& -8.8121, -2.7559, 70.3248
& -9.6028, -5.4145, 70.5540
& 2.7884 \\
\hline
\multirow{2}{*}{1} 
& Position:
& -0.1212, -1.2827, 1.2262
& 4.9537, -2.2304, 0.7702
& 4.9709, -2.2464, 0.7591
& 0.0259 \\
& Orientation: 
& -9.5741, -2.9049, 71.6770
& -10.4049, -7.8381, 23.8293
& -9.7747, -8.5313, 24.0012
& 0.9363 \\
\hline
\multirow{2}{*}{2} 
& Position:
& 4.9582, -2.2542, 0.7562
& 0.9875, -0.6979, 0.8351
& 0.9979, -0.7218, 0.8349
& 0.0260 \\
& Orientation: 
& -8.8006, -8.7949, 25.9951
& -14.1062, 9.9981, 43.3958
& -16.1230, 9.9351, 45.5272
& 2.6733 \\
\hline
\multirow{2}{*}{3} 
& Position:
& 0.9885, -0.7285, 0.8323
& -0.2504, -0.5382, 0.7834
& -0.2590, -0.5208, 0.7731
& 0.0220 \\
& Orientation: 
& -15.5100, 10.1585, 46.2320
& 9.2304, -3.7758, 72.6969
& 10.7258, -7.1104, 75.1778
& 4.3358 \\
\hline
\multirow{2}{*}{4} 
& Position:
& -0.2734, -0.5429, 0.7658
& 2.3408, -1.5267, 0.9428
& 2.3278, -1.5120, 0.9464
& 0.0200 \\
& Orientation: 
& 11.4821, -7.4198, 75.1721
& 5.7869, 11.2529, 65.9417
& 7.2479, 13.6307, 69.0070
& 4.3660 \\
\hline
\multirow{2}{*}{5} 
& Position:
& 2.3254, -1.5146, 0.9465
& 1.2252, -2.8447, 0.8277
& 1.1981, -2.7954, 0.8325
& 0.0565 \\
& Orientation: 
& 7.1448, 13.6192, 69.0242
& 5.3572, -8.3308, 85.1186
& 5.9588, -13.9744, 88.3386
& 6.4663 \\
\hline
\end{tabular}
\caption{Sequential pose manipulation results demonstrating student policy generalization across diverse pose configurations. For each experiment, the table presents the initial pose (derived from the previous experiment's final position), the randomly generated target pose, the measured final pose achieved by the student policy, and the corresponding tracking error. Position coordinates are given in meters [x, y, z], orientation angles in degrees [roll, pitch, yaw], with position errors and orientation errors in the same units respectively.}
\label{tab:generalize_poses}
\end{table*}

\subsubsection{Conclusion}
From these experiments, we deduce the following:
\begin{itemize}
    \item Our method has sufficient capacity to learn very different kinds of paths that the UAVs may need to follow for complex trajectories. This is evidenced by the \textit{SP (Both)} being able to generate very different trajectories required for zero-sideslip trajectories. Some illustrations are shown in Figure \ref{fig:tight_turns}
    \item Our method is not able to generalize to trajectories that lie outside the domain of the trajectories in the training dataset. This is especially true for trajectories involving novel sequences of orientations. The centralised teacher policy could follow all trajectories tested without any fine-tuning. In contrast, our method needs to be trained from demonstrations on similar trajectories for optimal performance.
    \item Our solution is limited by the simulation environment - it is not possible to conduct thousands of simulations parallelly on Gazebo. Other simulators purpose built for reinforcement learning applications allow this. A model trained on tens of thousands of random trajectories might be able to generalise better to novel trajectories.
\end{itemize}

\subsection{Robustness}
In this section we test our method's ability to handle mismatch in model dynamics and communication delays - scenarios it may encounter in the real world. As a baseline, we take the Student Policy (Both) that was trained to navigate both constant psi and zero side-slip trajectories. We test this policy, without any find tuning, on the Square zero-side slip trajectory in multiple scenarios in Gazebo simulation. These scenarios are described below:
\begin{table*}[h]
\centering
\small
\begin{tabular}{|l|l|r|r|}
\hline
\textbf{Experiment} & \textbf{Student Policy} & \textbf{RMSE Position (m)} & \textbf{RMSE Orientation ($^\circ$)} \\
\hline
No perturbation & SP (Both) & 0.181 & 13.107 \\
With 0.6 Kg Ball & SP (Both) & 0.271 & 17.795 \\
Load with double mass & SP (Both) & 0.334 & 44.469 \\
Attach point \& Rope mismatch & SP (Both) & 0.289 & 47.562 \\
Communication delay & SP (Both) & 0.178 & 12.586 \\
\hline
\end{tabular}
\caption{RMSE for position and orientation under different environment perturbations.}
\label{tab:rmse_experiments}
\end{table*}

\begin{figure*}[htbp]
    \centering
    \begin{subfigure}[b]{0.48\textwidth}
        \centering
        \includegraphics[width=\linewidth]{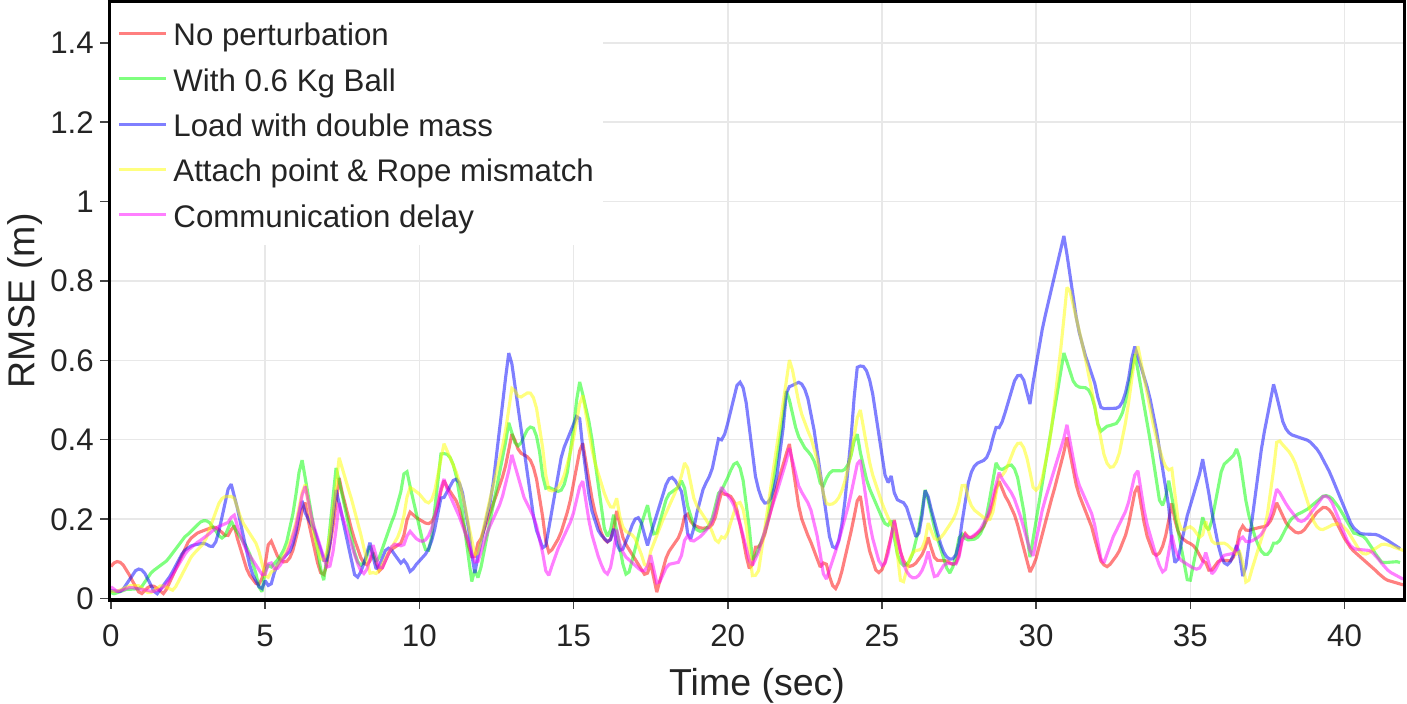}
        \caption{Position tracking error signal}
        \label{fig:robust_posn}
    \end{subfigure}
    \hfill
    \begin{subfigure}[b]{0.48\textwidth}
        \centering
        \includegraphics[width=\linewidth]{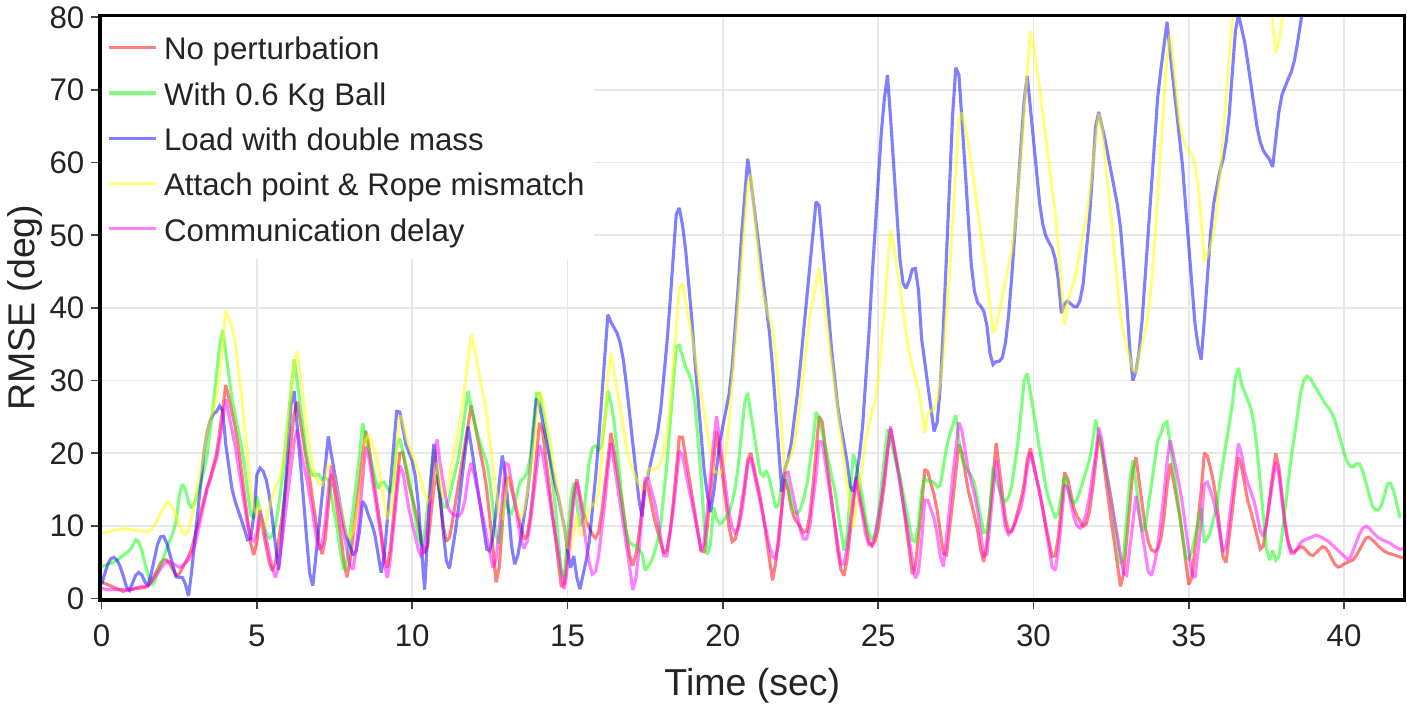}
        \caption{Attitude tracking error signal}
        \label{fig:robust_att}
    \end{subfigure}

    \vspace{0.5cm}
    
    \begin{subfigure}[b]{0.48\textwidth}
        \centering
        \includegraphics[width=\linewidth]{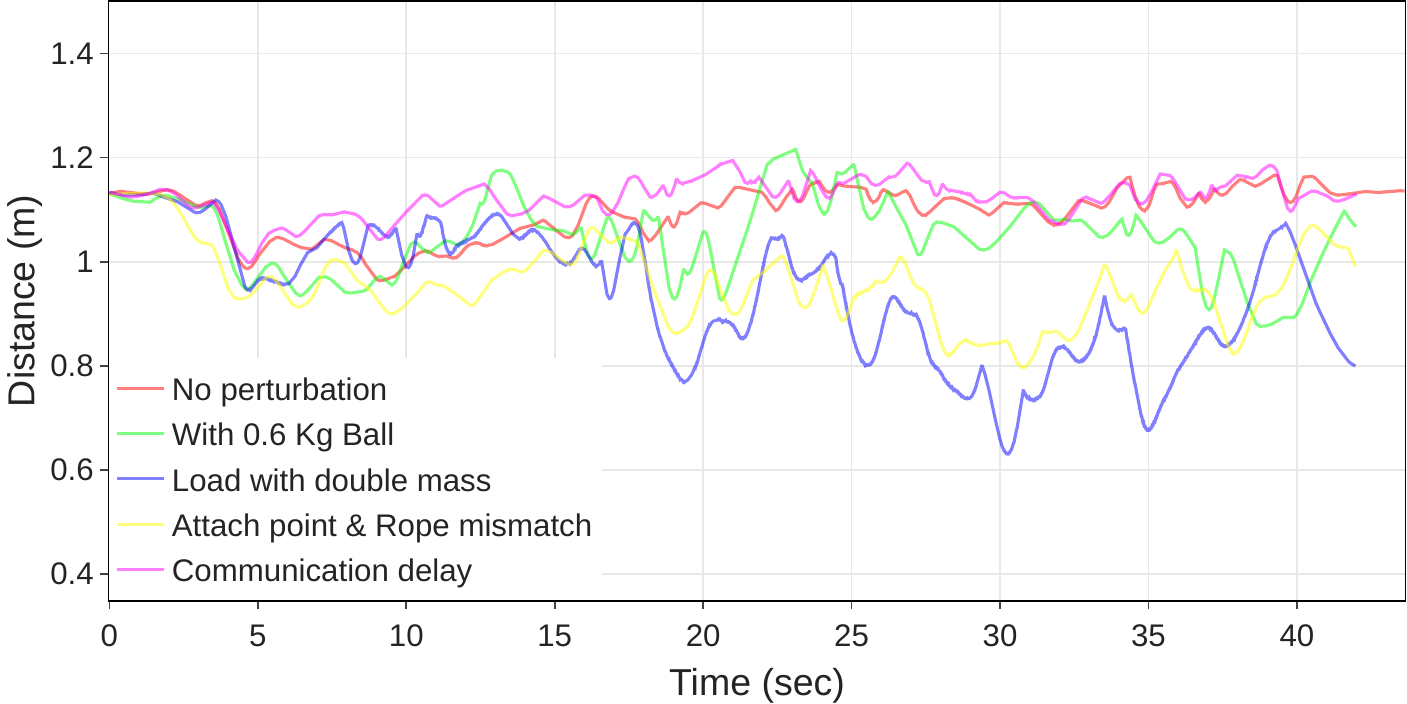}
        \caption{Minimum inter-agent distances}
        \label{fig:robust_dronedist}
    \end{subfigure}

    \caption{Performance signals of Student Policy (Both) to measure robustness to various environmental disturbances.}
    \label{fig:robustness}
\end{figure*}

\begin{figure}[h!]
    \centering
    \includegraphics[width=0.9\linewidth]{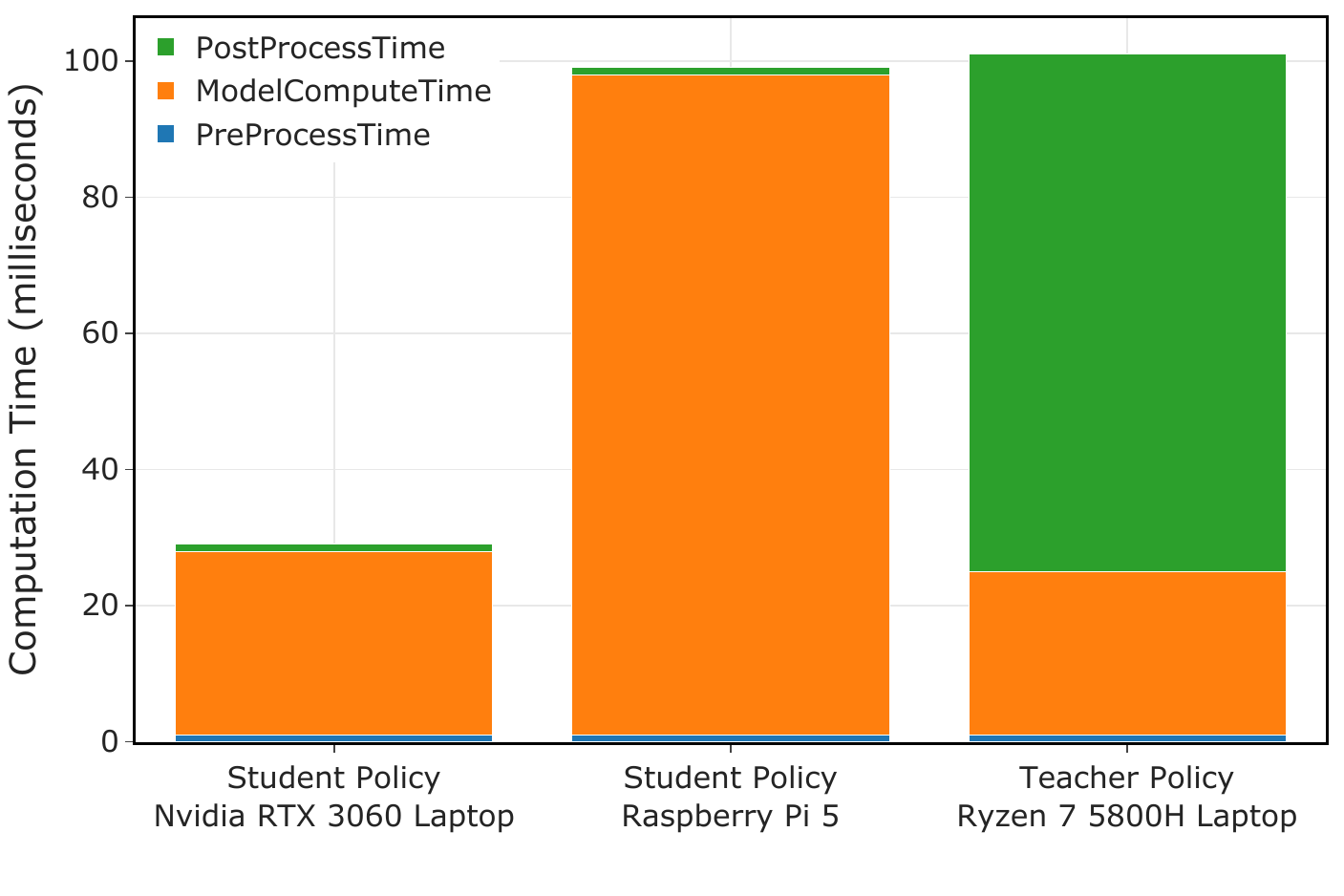}
    \caption{Computation time breakdown for teacher and student policies. Total time is divided into three components: (1) ModelComputeTime, representing the time required for inference using either the ML model or the OCP solver; (2) Pre-Processing; and (3) Post-Processing, which involve organizing input/output data for compatibility with the respective algorithms.}
    \label{fig:computaion_time}
\end{figure}

\begin{itemize}
\item \textbf{Sloshing Load:} A 0.6 kg ball is placed inside the 1.4 kg payload lifted by the UAVs. This ball is free to move relative to the payload structure, significantly altering the inertial characteristics of the system during flight. The disturbance becomes especially severe when the ball shifts rapidly, impacting the boundaries of the payload basket. Although this introduces unpredictable dynamics, our proposed method demonstrates robustness to these variations, albeit with some degradation in pose tracking accuracy.
\item \textbf{Mass Variation:} The total payload mass is increased from 1.4 kg to 2.8 kg. The resulting higher inertia makes trajectory tracking particularly challenging, especially during sharp turns. Additionally, the increased cable tensions often disrupt the formation, occasionally bringing the UAVs close to each other. Despite these challenges, our approach remains functional, though with a significant reduction in tracking performance.
\item \textbf{Attachment Point and Rope Length Variations:} All attachment points on the payload are shifted by a few centimetres. Furthermore, asymmetry is introduced by adjusting the lengths of two out of three suspension ropes: one is shortened by 3.5 cm while another is lengthened by the same amount. Our method exhibits high sensitivity to such changes in attachment geometry, with even small deviations leading to substantial performance degradation in orientation tracking.
\item \textbf{Communication Delay:} To evaluate robustness to communication latency, we simulated delays in transmitting the student policy’s predictions to the Agilicious flight stack. Specifically, 20\% of predictions were dropped, and for the remaining messages, a random delay of up to 30 ms was introduced. Due to our approach of generating receding horizon trajectories with a 2-second time window, degradation in communication did not adversely impact overall system performance.
\end{itemize}

\subsection{Computation Time}
Real-time inference is possible on the Raspberry Pi 5 CPU and takes 97 milliseconds. As the policy generates trajectories at 10~Hz, the computation time is just within the available computation budget. However, during experiments, we realized that running both the Agilicious flight control stack and PINN inference on the Raspberry Pi leads to a degradation in the inner control loop update rate. Therefore, we eventually deployed PINNs on a laptop and transmitted trajectories to each UAV through Wi-Fi. The average inference time of the student policy on an NVIDIA RTX 3060 Laptop GPU is 27 milliseconds. The computation time onboard the Raspberry Pi can be drastically reduced by using an AI accelerator along with optimised inference of the ML model.

In comparison, the teacher policy requires an average computation time of 101 milliseconds (for 3 UAVs) running on a Ryzen 7 5700H CPU.
Of this, the nonlinear model predictive control (NMPC) solver accounts for 24 milliseconds, while post-processing operations—such as computing executable drone trajectories from the solver outputs—consume an additional 76 milliseconds. An optimised C++ implementation can considerably reduce the overall computation time of the teacher policy.

\begin{table*}[htbp]
\centering
\small
\begin{tabular}{|l|c|c|c|c|}
\hline
\textbf{Name} & \textbf{Rounds} & \textbf{Data Collection Time} & \textbf{Training Time} & \textbf{Total Time} \\
\hline
SP (Real) & 25 & 93.89 & 25.35 & 119.24 \\
SP (Constant orientation) & 30 & 97.62 & 104.72 & 202.34 \\
SP (Zero sideslip) & 30 & 88.76 & 98.23 & 186.99 \\
SP (Both) & 60 & 178.67 & 388.37 & 567.04 \\
\hline
\end{tabular}
\caption{Data collection, training, and total time for different SP configurations. All values are in minutes. SP (Real) was the policy deployed in real world experiments. }
\label{tab:sp_times}
\end{table*}

Unlike the student policies, the teacher's computational complexity increases exponentially with the number of UAVs. 
In contrast, the student policy, due to its decentralized and partially observable architecture, exhibits the potential for scalability. By enabling each UAV to independently predict its own trajectory, the framework could support horizontal scaling, making it feasible to accommodate a large number of UAVs without any increase in computation time.

\subsection{Training Time}
To better understand the time requirements of our learning pipeline, we disaggregate the overall training time into two primary components: data collection and model training. In each round of DAgger, demonstration data is collected from a single episode. This is followed by training the machine learning model on the cumulative dataset of demonstrations collected until now. Each episode yields a 40-second trajectory, and given that our simulator operates at 0.25× real-time, collecting one trajectory requires approximately 160 seconds. In contrast, training the model is relatively quick in the early stages but becomes progressively slower as the dataset grows. Table \ref{tab:sp_times} illustrates the training durations for the various policies introduced in the previous sections.

Our use of imitation learning enables efficient policy learning, significantly reducing training time compared to reinforcement learning techniques. By leveraging the full prediction horizon of the expert (teacher) policy, we enhance sample efficiency and accelerate convergence, even when constrained by a slow simulator. Despite these limitations, our method demonstrates robust performance. 

Integrating faster, parallelisable simulation environments such as MuJoCo or Isaac Sim would facilitate quicker data collection across a broader set of trajectories. This, in turn, could enable the learning of more generalizable policies capable of adapting to diverse trajectories.

\subsection{Human-Robot Interface} \label{sec:hri}
The human-robot interface for this system consists of 3 key components:
\begin{itemize}
    \item \textbf{GUI:} A graphical-user interface allows the operator to check the status off all UAVs, including their state (position, velocity), their battery's charge level and the status of the flight controller. This GUI also allows the controller to choose the desired trajectory along which the payload must be transported by the UAVs. Figure \ref{fig:gui} shows the user interface.
    \item \textbf{Controller:} The system can be operated using a handheld controller connected to the ground station via Bluetooth, enabling the operator to issue high-level commands. The controller is equipped with a dead-man switch mechanism: releasing a designated trigger immediately powers down all UAVs, thereby enhancing safety and minimizing the risk of hardware damage. The operator can initiate high-level actions such as activating the Agilicious flight stack, taking off, landing, commanding the UAVs to hover, and shutting them down. Additionally, the controller allows the operator to start the payload transport trajectory. Upon initiation, the system’s policy begins generating the corresponding UAV trajectories, which are displayed to the operator for verification. Only after manual confirmation can the UAVs begin following the predicted trajectories. The controller’s button mapping is illustrated in Figure \ref{fig:xbox}. 
    \item \textbf{RViZ:} This interface allows the operator to see a visual of the state of the load and all UAVs. This interface also allows the operator to see the trajectories generated by the teacher or the student policies. This feature is particularly useful as it allows the operator to visually confirm the generated trajectories make sense before enabling UAVs to fly these trajectories thereby giving the operator meaningful control over the system.
\end{itemize}

\subsection{Safety Module} \label{sec:safety_module}
The safety module is designed to continuously observe the state of the UAVs and the trajectories generated by the policies. Based on this data, the safety node can take measures to prevent collisions or damage to the UAVs.

\begin{figure*}[htbp]
    \centering
    \begin{subfigure}[b]{\textwidth}
        \centering
        \includegraphics[width=\textwidth]{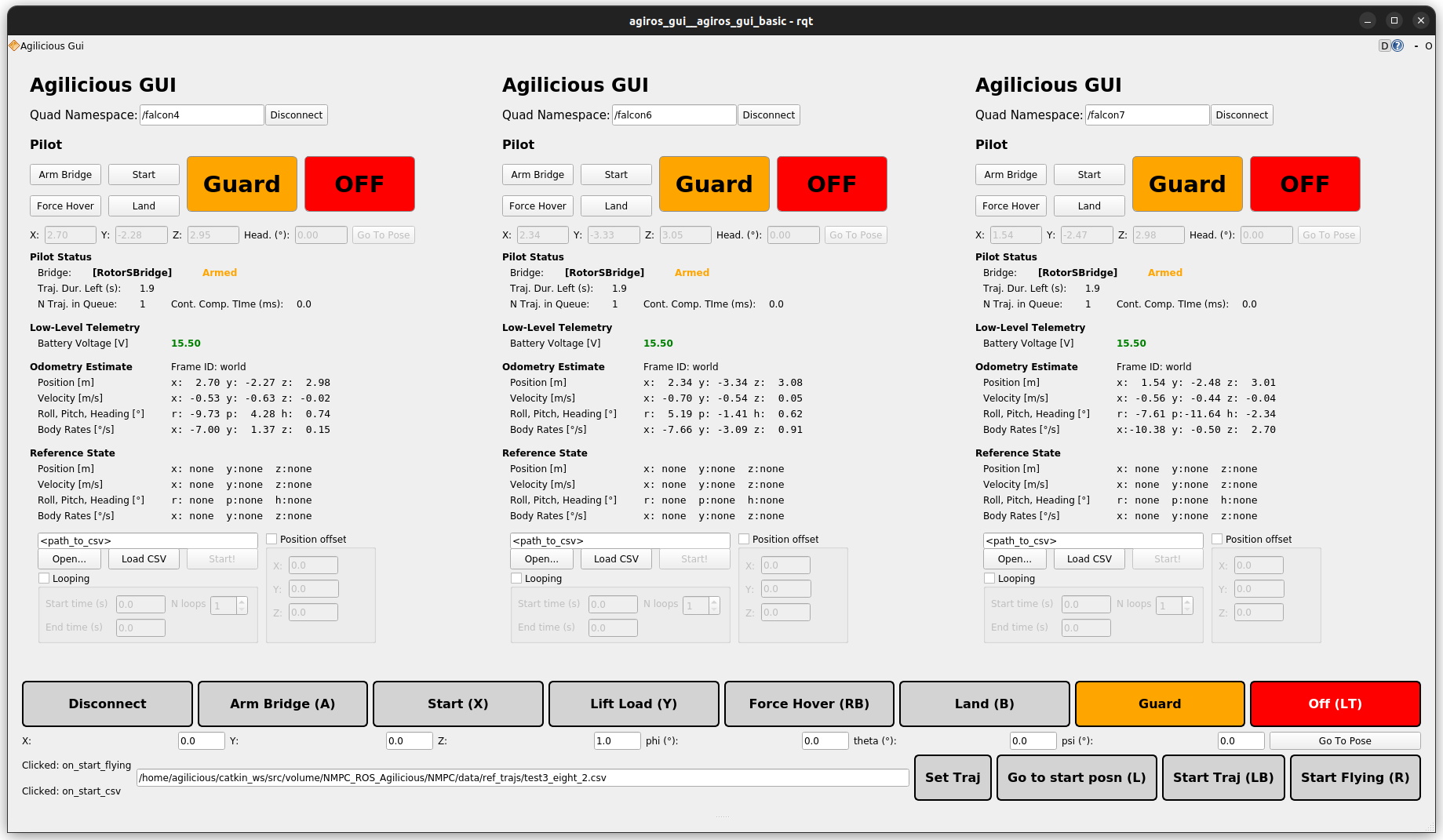}
        \caption{User interface for controlling the system, extending the original Agilicious GUI for the multi-UAV carrying system. Users can select the desired trajectory in the menu at the bottom.}
        \label{fig:gui}
    \end{subfigure}
    \begin{subfigure}[b]{0.48\textwidth}
        \centering
        \includegraphics[width=\textwidth]{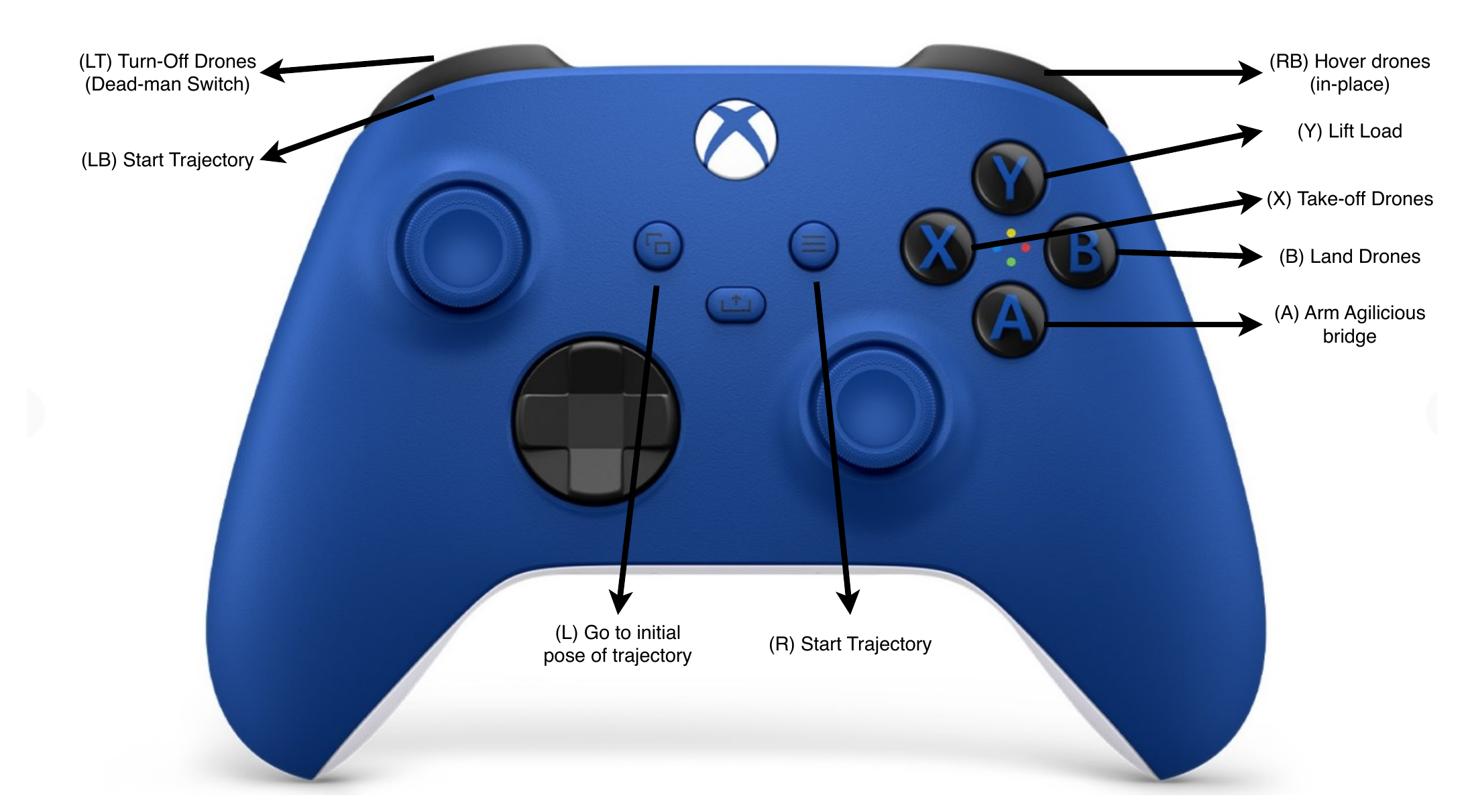}
        \caption{Xbox controller button mappings. The controller includes a deadman switch that, when released, turns off all UAVs immediately.}
        \label{fig:xbox}
    \end{subfigure}
    \hfill
    \begin{subfigure}[b]{0.48\textwidth}
        \centering
        \includegraphics[width=\textwidth]{images/square_no_yaw.drawio.png}
        \caption{RViZ interface showing the current state of the system, including the load, UAVs, and predicted trajectories (red, green, blue lines).}
        \label{fig:rviz}
    \end{subfigure}
    \caption{Human-robot interface: (a) GUI for system control and trajectory selection; (b) Xbox controller for issuing high-level commands; (c) RViZ visualisation of UAV and load states and trajectories.}
    \label{fig:hri}
\end{figure*}

The safety node has the following features:
\begin{itemize}
    \item \textbf{Position Bounds:} The safety node continuously monitors the pose of all UAVs and turns them off if they go beyond the safe boundaries of the Arena.
    \item \textbf{Velocity and Acceleration Bounds:} The safety node monitors the twist and acceleration of all UAVs and forces them to hover in place if these values are more than the maximum permissible limit.
    \item \textbf{Distance Bounds:} Using the pose measurements of all UAVs, the safety node monitors whether any two drones are coming too close to each other. If UAVs come closer than the minimum permissible distance, the UAVs are turned off.
    \item \textbf{Future Reference Check:} The safety node also monitors the reference trajectories of the UAVs to ensure any of the aforementioned bounds are not violated. In case of violations, the UAVs are forced to hover in place.
\end{itemize}

For all instances where the safety node intervenes, a message showing the error is also displayed on the GUI shown in Figure \ref{fig:gui}.

\end{document}